\documentclass[final,3p,times,twocolumn]{elsarticle}

\usepackage[none]{hyphenat}
\usepackage{microtype}

\usepackage{amssymb}
\usepackage{amsmath}

\usepackage{array}
\usepackage{booktabs}
\usepackage{placeins}
\usepackage{enumitem}

\usepackage{lineno}
\usepackage{tikz}
\usepackage{tabularx}
\usepackage{float}
\usepackage{amssymb}

\usepackage{hyperref}

\usetikzlibrary{shapes.geometric, arrows, positioning, fit}

\definecolor{softgreen}{RGB}{173, 235, 173}
\definecolor{softteal}{RGB}{102, 204, 204}
\definecolor{softorange}{RGB}{255, 204, 153}
\definecolor{softpurple}{RGB}{204, 153, 204}

\begin{document}

\begin{frontmatter}



\title{Adaptive Visual Perception for Robotic Construction Process: A Multi-Robot Coordination Framework}


\author[label1]{Jia Xu} 
\author[label1]{Manish Dixit} 
\author[label1]{Xi Wang \corref{cor1}} 

\affiliation[label1]{organization={Department of Construction Science, Texas A\&M University},
            addressline={3337 TAMU}, 
            city={College Station},
            postcode={77843}, 
            state={TX},
            country={United States}}
\cortext[cor1]{Corresponding author}

\begin{abstract}
Construction robots operate in unstructured construction sites, where effective visual perception is crucial for ensuring safe and seamless operations. However, construction robots often handle large elements and perform tasks across expansive areas, resulting in occluded views from onboard cameras and necessitating the use of multiple environmental cameras to capture the large task space. This study proposes a multi-robot coordination framework in which a team of supervising robots equipped with cameras adaptively adjust their poses to visually perceive the operation of the primary construction robot and its surrounding environment. A viewpoint selection method is proposed to determine each supervising robot's camera viewpoint, optimizing visual coverage and proximity while considering the visibility of the upcoming construction robot operation. A case study on prefabricated wooden frame installation demonstrates the system's feasibility, and further experiments are conducted to validate the performance and robustness of the proposed viewpoint selection method across various settings. This research advances visual perception of robotic construction processes and paves the way for integrating computer vision techniques to enable real-time adaption and responsiveness. Such advancements contribute to the safe and efficient operation of construction robots in inherently unstructured construction sites.

\end{abstract}




\end{frontmatter}



\section{Introduction}

Robots, capable of performing well-defined tasks with high precision, hold significant promise for enhancing productivity and reducing the physical demands on workers in various industries, including construction \cite{daviladelgado2019}. Robotic arms, widely used in manufacturing due to their flexibility and precision, have been investigated by researchers for various construction applications, such as drywall installation \cite{wang2021, wang2024}, welding \cite{zhang2022}, pipe maintenance \cite{zhu2022robot}, wood frame assembly \cite{wongchong2022}, and ceiling installation \cite{liang2020, yu2024a}. While these research efforts highlight the potential of using robotic arms for construction tasks, the unstructured nature of construction sites poses unique challenges for the adoption of robots \cite{xu2020site}.

In controlled manufacturing environments, robotic arms are typically stationary, operating in designated, enclosed spaces and repetitively performing predefined tasks as products are delivered to their work area via conveyors. In contrast, on-site construction robots need to move around to perform tasks at varying locations, making it impractical to assign them fixed, separate workspaces \cite{gharbia2020}, which necessitates them to be able to work in a shared, open workspace. On a construction site, this means the potential existence of building structures, stacked construction materials, and even moving human workers and machinery surrounding the robot. The large size of the construction materials these robots handle (e.g., drywall panels) further adds to the potential of unintended contact. Therefore, it is crucial to maintain keen awareness of the surroundings of field construction robots, the robots themselves, and the objects they handle \cite{helm2012, oyediran2024human}. Visual perception is an important source of this awareness \cite{yang2021}.

Robotic visual perception primarily relies on data from vision sensors such as depth cameras and RGB cameras, mounted on the robot's body or installed in surrounding environments. Visual data obtained from onboard vision sensors can be processed using various computer vision techniques, enabling the robot to understand its environment \cite{fan2022}, determine its position \cite{liu2024}, locate target objects \cite{feng2015, rogeau2020} and control assembly quality \cite{prezas2022}. However, the perception of onboard sensors often faces limitations due to occlusion from the robotic arm itself and the large elements it handles. Additionally, as the robot moves, the sensors shift position and orientation, further restricting the Field Of View (FOV) needed to perceive the robot surrounding environment or target areas effectively.

In addition to onboard vision sensors, environmental cameras configured around the task area have been widely used to supplement onboard sensors by providing a broader visual perspective, crucial for mitigating potential safety risks in human-robot shared environments and for tasks such as collision avoidance \cite{kim2020proximity, mohammed2017} and workspace monitoring \cite{kang2019,lopes2019}. Visual feedback from multiple static cameras has also been used to enhance the teleoperation of equipment such as excavators \cite{chae2024, kamezaki2024}. Despite these advantages, environmental cameras face significant challenges in large and evolving construction sites. The core construction work areas are frequently shifting and constantly evolving based on the construction schedule and process. Frequent repositioning is often required to cover different areas or adapt to up-to-date layouts, a process made more complex and time-consuming by the partially built infrastructure typically found on active construction sites. Moreover, these environmental cameras often lack the adaptability needed to account for the varying poses of the construction robot along the task execution process and struggle with the occlusions caused by the robot's body and the object manipulated by the arm, which is moving with the arm along its operational trajectory \cite{luo2021, zhu2020}. These limitations underscore the need for a more adaptable visual perception system for robotics construction processes. 

This paper proposes a novel approach for enhancing the visual perception of robotic construction processes. We present a multi-robot coordination framework where a group of ``supervising robot'' provides visual support to the primary robot executing construction tasks. These supervising robots dynamically adjust their onboard camera poses to perceive the primary robot's motions, its surrounding environment, and the object it manipulates. This framework offers adaptive visual perception that can accommodate various robot models, tasks, and trajectories. The proposed framework comprises four main modules: a Building Information Model (BIM) repository to guide robotic construction operations, a construction robot motion planning and execution module for the primary construction robot, a camera viewpoint selection module to determine camera poses for supervising robots, and a supervising robot control module for positioning the supervising robots to achieve target camera poses. By integrating these modules, our proposed framework has the potential to enhance the efficiency and safety of construction robotic operations in unstructured construction sites through enhanced visual monitoring with coordinated supervising robots. To verify this approach, a case study focused on prefabricated wooden frame installation is conducted. Furthermore, a series of experiments are performed to evaluate the proposed viewpoint selection methods across varying layouts, space constraints, and material positioning. These experiments aim to demonstrate the effectiveness and adaptability of the proposed viewpoint selection method across diverse construction scenarios. 

The proposed framework enables supervising robots to adaptively deliver essential and high-quality visual data, which serves as the foundation for construction robotic systems to interpret the surroundings and respond accordingly in unstructured construction sites. This visual data supports downstream tasks such as collision avoidance \cite{nascimento2020collision, fabrizio2016real, saverianoDistanceBasedDynamical2014}, human intention recognition \cite{cai2024fedhip, liu2024intention, wang2024b}, and vision-guided assembly \cite{feng2015,liu2024automatic}. Furthermore, it has the potential to be further integrated with rapidly evolving image segmentation models (e.g., SAM \cite{kirillov2023}), vision foundation models like Dinov2 \cite{oquab2024}, and vision-language foundation models such as CLIP \cite{radfordLearningTransferableVisual2021} to empower a more intelligent and decision-capable construction robot. While the detailed implementation of these applications is beyond the scope of this paper, the method has the potential to significantly advance the integration of robots in construction by enabling the context awareness necessary for their safe and adaptive operation in dynamic construction environments and seamless coexistence with human workers.

The remainder of this paper is structured as follows. Section~\ref{related-works} reviews related works on visual data collection and multi-robot systems in construction. Section~\ref{system-framework} presents the proposed multi-robot coordination framework, with the viewpoint selection method detailed in Section~\ref{optimal-view-selection}. A case study on prefabricated wooden frame installation is provided in Section~\ref{case-study}, and Section~\ref{experimental-validation} presents experimental assessments under varying settings. Lastly, Section~\ref{discussion} discusses the implications, limitations, and potential future work, followed by Section~\ref{conclusion} concluding the paper.

\section{Related Works}\label{related-works}
\subsection{Visual Data Collection}

Visual perception data from onboard or environmental vision sensors is critical for the effective and safe operation of robots in complex environments, such as construction sites. This data can be interpreted using various computer vision techniques to support multiple applications. For example, images obtained can be used for object detection and recognition \cite{zhao2023bim}, image semantic segmentation \cite{yang2024enhanced, yang2023cost} and object pose estimation \cite{duVisionbasedRoboticGrasping2021, tremblay2018deep}, enabling robots to locate and interact with target objects. In construction settings, such data provides guidance for essential tasks like grasping materials \cite{asadi2021} and component installation \cite{qin2016}. Additionally, the integration of depth camera data supports collision avoidance \cite{kim2020proximity, mohammed2017}, allowing robots to safely interact with the surroundings. In scenarios involving human-robot coexistence or collaboration, external visual data helps robots detect human presence \cite{lopes2019}, recognize human activities \cite{amani2024adaptive, sherafat2020automated}, interpret human gestures \cite{yoon2024laserdex,liu2018}, infer human intentions \cite{cai2024fedhip, cai2023multi}, and predict human actions \cite{liu2024intention, xia2022human}, facilitating smooth and safe interactions between humans and robots \cite{robinson2023}.

Despite the broad applications, the reliability and accuracy of these computer vision-based systems are heavily dependent on the quality and completeness of the input data \cite{dong2022, sun2019}. Factors such as occlusions, blind spots, or suboptimal angles can greatly impact the performance of these algorithms. Therefore, it is crucial to strategically plan the sensor views to minimize occlusions and improve critical features' visibility for better downstream processing and interpretation. Environmental camera placement optimization and robot active vision have been extensively investigated to address these challenges.

Camera sensors placed near the robot's work area can provide detailed visual information about the robot and its surroundings. However, visual perception from a single camera is restricted to its FOV, offering only limited information. To overcome this limitation, one effective way is to deploy multiple sensors strategically to obtain more comprehensive visual data. Studies on optimizing camera sensor placement emphasize maximizing area coverage \cite{altahir2017, horster2006, joshi2009}, which are formulated as a combinatorial optimization problem that incorporates various constraints such as camera configurations, environment layouts, and specific user requirements \cite{liu2016}. In the construction context, camera placement research primarily focuses on optimizing positions for construction process monitoring \cite{houng2024, kim2018a, chen2021} and building surveillance \cite{yang2018, albahri2016, chen2023smart}. These studies typically aim to maximize site or building coverage while minimizing installation costs \cite{kim2018a, kimSystematicCameraPlacement2019}. However, these studies lack consideration of the evolving nature of construction sites. To address this challenge, studies integrate BIM and construction schedules to optimize the camera placement, maximizing overall coverage while considering the site layout changes at different construction phases \cite{chen2021, houng2024}.

However, these studies primarily focus on changes in static occlusions (e.g., those caused by building structures) during different construction phases and coverage of fixed areas but do not account for dynamic occlusion issues. In robotic construction tasks, elements being handled or the heavy-duty industrial robot arms themselves can cause dynamic occlusions. To address this, Zhu et al. \cite{zhu2020} use multi-objective optimization to minimize occlusions across various industrial robot poses. However, it relies on randomly generated robot poses instead of considering specific task-performing processes. Recognizing the need to account for task-specific requirements, researchers explore camera placement tailored to the movement patterns of specific targets. For example, Bodor et al. \cite{bodor2007} optimize camera placement based on the distribution of the predicted human movement paths. While effective in confined spaces like rooms, these methods are less applicable to large-scale, dynamic environments like construction sites where distributions of construction robots are relatively unpredictable. Their motions are task-specific, depending on various task objectives and specifications, complicating camera placement.

In addition to environmental sensors, another important visual data source is the robot's onboard sensors. To enhance the robot's effectiveness in obtaining useful visual data, active robot vision has gained significant attention \cite{chen2011}. Instead of simply using the passively obtained visual input, active vision methods control the robot to actively obtain better visual input \cite{queralta2020}. For example, given the limited information from a single viewpoint, view planning has been used to determine a set of viewpoints or the next best view for robots to progressively cover a scene or object (scene{/}object reconstruction) \cite{kaba2016, vasquez-gomez2014, zeng2020}, detect target objects (object recognition) \cite{ammirato2017} or perform non-contact pose estimation \cite{atanasovNonmyopicViewPlanning2014, huViewPlanningObject2022}. Active vision is promising to improve construction robots' perception by enabling the collection of more comprehensive visual information through adaptive camera adjustments. However, for robotic arms performing construction tasks, simultaneously manipulating the camera and executing the primary task is often impractical. Adjusting the camera pose when they are holding heavy objects increases the risks and is not efficient. Therefore, we propose introducing flexible external robots to perceive the robotic construction process with a strategically planned vision.

\subsection{Multi-Robot System}\label{multi-robot-collaboration}

A Multi-Robot System (MRS) enables multiple robots to work together cooperatively to accomplish complex tasks \cite{gautam2012}. This approach has gained extensive attention in the exploration of large-scale unknown areas \cite{burgard2000, alitappeh2022}, search and rescue applications \cite{queralta2020}, and large-scale scene reconstruction \cite{dong2019}. In these applications, multiple robots can simultaneously address different aspects of a task, substantially reducing the execution time. In construction, MRS has also gained more and more attention for its potential to enhance efficiency, safety, and adaptability in various construction processes \cite{krizmancic2020}. Zhang et al. \cite{zhang2018} utilizes a team of mobile robots for large-scale 3D printing, demonstrating the potential for these systems to operate collaboratively on-site. In another multi-robot 3D printing case, the YouWasp platform was developed, which employs a team of mobile extruder robots that autonomously perform material deposition, leveraging collision-aware printing and task allocation strategies to improve scalability and adaptability on construction sites \cite{sustarevas2019}. Apart from multi-mobile robot 3D printing, multi-robot systems have been explored in cooperatively assembly tasks, including bricklaying \cite{elkhapery2023}, and timber framing \cite{adel2024}. A feedback-driven adaptive multi-robot system for timber construction is proposed, utilizing pose-based and topology-based methods to handle material uncertainties, improve accuracy, and reduce deviations through real-time feedback \cite{adel2024}. Additionally, optimization of team composition and task allocation \cite{pan2024, ye2024} in construction has also been explored to improve coordination efficiency. Optimization algorithms and strategies have been explored to determine the optimal number and types of robots for specific construction tasks and how to distribute tasks among available robots efficiently.

In addition to collaborating directly on task performing, another form of collaboration involves an assistant robot offering a view of the task being carried out by the primary robot \cite{dufek2021, xiao2021}. Assistant robots for visual support have gained attention in teleoperation applications. Researchers have explored using unmanned aerial vehicles  \cite{senft2022, xiao2021}, mobile robots \cite{maeyama2016}, or robotic arms \cite{nicolis2018, rakita2018} as assistant robots to adaptively adjust their camera's viewpoint, thereby improving teleoperation support. For example, Rakita et al. \cite{rakita2018} employs a secondary camera-in-hand robot arm that automatically adjusts to provide the best possible view of end effector for remote teleoperation. Motion prediction is used to anticipate where the end effector will be in the near future, allowing the camera to adjust its viewpoint proactively. The system uses a weighted sum objective function combining multiple criteria, including visual target focus, occlusion avoidance, and smooth camera motion, to decide the view.

In these teleoperation applications, which manipulate small objects, ensuring a good view of the robot's end effector is generally adequate to accomplish tasks effectively. However, in construction settings, the scale of elements involved in assembly tasks is considerably larger. This not only increases the difficulty of achieving comprehensive visual coverage of the element but also leads to substantial visual obstruction. This presents a unique challenge that goes beyond simply focusing on the end effector. Therefore, developing a method that efficiently integrates assistant robots to gather useful visual data for robot construction tasks is crucial. Limited research has been done in construction to provide visual support with assistant robots. Kamezaki et al. \cite{kamezaki2024} developed a multi-flying camera system using drones to provide an overhead view, back-manipulator view, side-manipulator view, and cab view to excavator teleoperators. The cameras dynamically follow the excavator while maintaining fixed relative positions and angles. However, the fixed relative position lacks the ability to optimize visibility for different operations and varying space constraints. It may also result in the excavator itself or environmental obstacles frequently blocking viewpoints.

To address the above-mentioned research gaps, a new workflow is needed and this workflow should embody the following characteristics. First, instead of relying on fixed environmental cameras for visual perception, assistant robots with active vision offer greater flexibility, allowing them to adapt to large task spaces for robotic construction. Secondly, rather than focusing on the observability of the preprogrammed movements, supervising robots should adapt to the upcoming construction robot motion, which varies from one operation to another. Additionally, unlike visual assistants in teleoperation, which focus on providing a clear view of the end effector, the assistant robot should also consider the large dimensions of the target (i.e., the construction elements) to maintain visibility. Therefore, the following section presents a multi-robot framework that enables assistant robots to adaptively perceive the robotic construction process.

\section{System Framework}\label{system-framework}

\subsection{Overview}\label{overview}

\begin{figure*}[h!]
  \centering
  \includegraphics[width=0.8\textwidth]{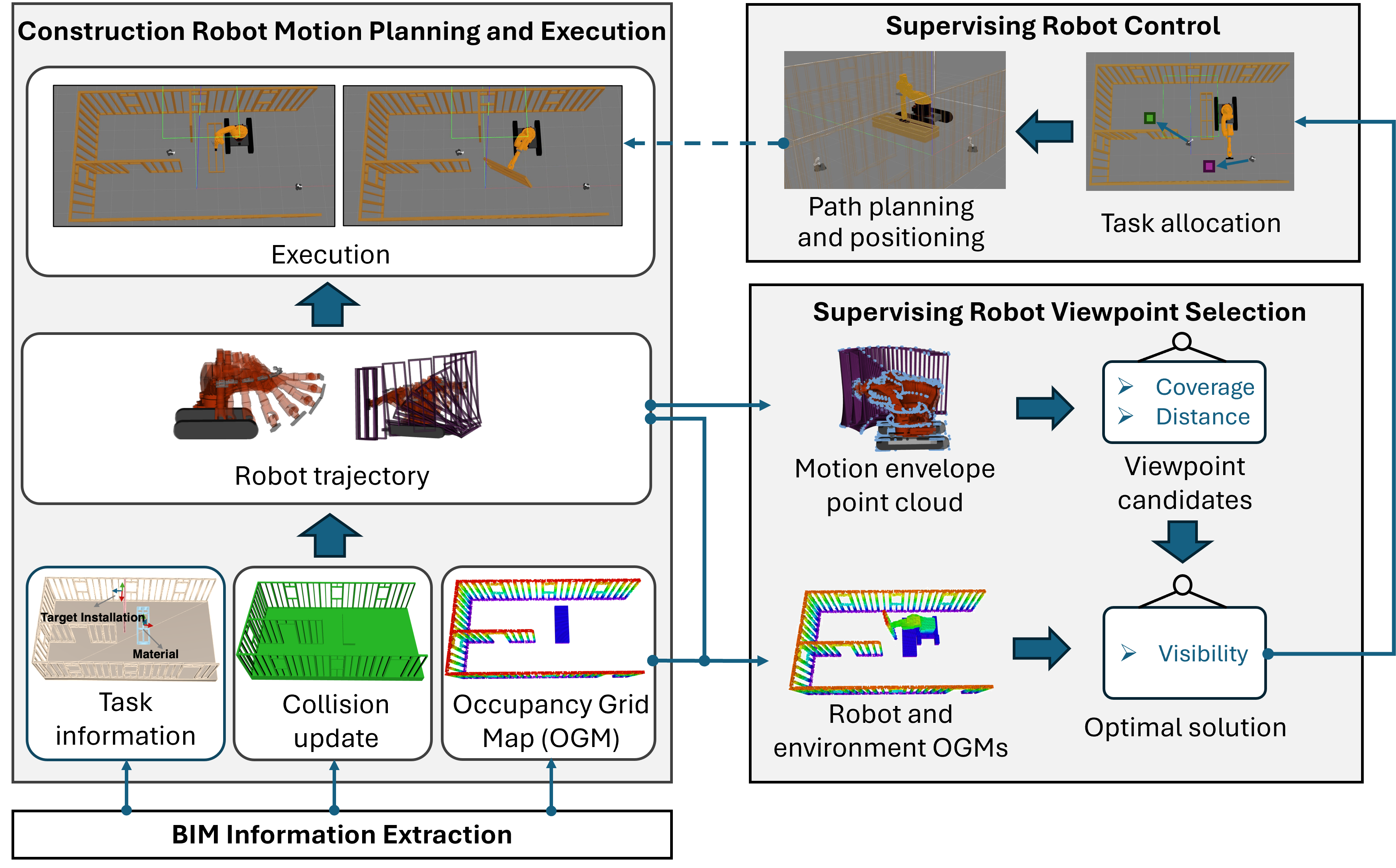}
  \caption{Overall framework of multi-robot coordination}
  \label{fig:overall-framework}
\end{figure*}

This study proposes a multi-robot framework that coordinates a heterogeneous robot team consisting of a construction robot and supervising robot(s) to achieve adaptive visual perception for upcoming robotic construction processes (Figure~\ref{fig:overall-framework}). While the construction robot performs its tasks, a team of supervising robots provides visual support by dynamically adjusting their cameras to determine viewpoints. This coordination enables the supervising robots to continuously capture the construction robot's task performance.

As shown in Figure~\ref{fig:overall-framework}, the primary framework consists of four main modules:

\textbf{1) BIM information extraction module}: This module is responsible for extracting and publishing relevant BIM data to guide robotic construction tasks. It provides crucial information about the surrounding environment, materials, and construction targets, enabling effective task and motion planning of the construction robot.

\textbf{2) Construction robot motion planning and execution control module:} This module processes the received BIM information to guide the construction robot's decision-making and execution processes. It transforms geometric and attribute data from the BIM into task objectives (e.g., material and installation target poses) and collision scenes, generates collision-free joint space trajectories, as well as controls robot execution. The module also converts BIM geometries into a 3D Occupancy Grid Map (OGM) for subsequent visibility evaluation.

\textbf{3) Viewpoint selection module:} This module is the core of the proposed multi-robot coordination processes, designed to determine viewpoints for the involved supervising robots to collectively capture the construction process. It analyzes the next-step trajectory of the construction robot and determines the viewpoint combinations for available supervising robots. These viewpoints collectively achieve comprehensive coverage of the construction process, minimize the distance to the targets, and maximize the visibility of target objects.

\textbf{4) Supervising robot control module:} Once the viewpoints are selected, this module generates and executes control commands for the supervising robots to position their cameras to achieve the selected viewpoints. It enables the coordination and movement of supervising robots.

By integrating these modules, the framework enables coordination between the construction and supervising robots for adaptive visual perception of the robotic construction process. The obtained visual information serves downstream computer vision-based applications for enhancing environmental awareness and enabling safe operation in unstructured construction sites. The following section introduces the technical approach that enables this proposed framework.

\subsection{BIM Information Extraction Module}\label{bim-information-extraction-module}

This study integrates BIM based on the framework proposed by Wang et al. \cite{wang2024}. We assume that current BIM is kept up-to-date using techniques such as robotic mapping \cite{kim2018slam, gan2024automated}, Scan-to-BIM \cite{chen2022rapid, wang2024omni}, and change detection\cite{chuang2023, meyer2022}. This ensures the BIM accurately reflects the current state, including as-built and un-built components, temporary structures, and materials. Based on Wang et al. \cite{wang2024}, Rhino is employed as the BIM platform, where three layers---representing as-built elements, material elements, and target elements scheduled for assembly---are integrated. BIM information is extracted with the COMPAS FAB package \cite{zotero-1128} and published to the other modules of the system through RosBridge \cite{crick2017rosbridge}. This enables ROS to interface with the BIM and allows the construction robot to access the necessary information for task and motion planning.

Table \ref{tab:bim-structure} outlines the three-layer BIM data structure and the specific attributes of the objects needed for the construction robot's task planning, detailed as follows: \textbf{(1) As-built Layer:} This layer contains previously built and other existing objects. Attributes for objects in this layer, including the component's name and mesh, provide the data needed to update the collision map for collision-free robot motion planning; \textbf{(2) Materials Layer:} This layer includes the construction material objects on site. Attributes for each object in this layer include its position, orientation, type, picking direction, and offset. The picking direction specifies how the robot should grasp the material. The offset, along with position and rotation, can be transformed to target picking poses for the construction robot end effector. This allows for precise robotic manipulation of construction materials; and \textbf{(3) Target Layer:} This layer contains objects scheduled for installation. The attributes of each object in this layer include its position and local normal vector relative to the world frame, type, and order in the work sequence. The robot end-effector pose for installation is calculated based on the target object's global position and local normal vector. The type specifies the material category, which helps select the corresponding material components, while the order indicates the installation sequence of the targets. The target layer guides the robot in placing components accurately and sequentially.


\begin{table}[h!]
\small
\centering
\caption{Object Attributes by Layer}
\label{tab:bim-structure}
\begin{tabular}{l l}
\toprule
\textbf{Layer Name} & \textbf{Object Attribute} \\
\midrule
As-built Layer & Name and Mesh \\
Material Layer & Name, Position, Rotation, \\ & Picking direction, Offset, and Type \\
Target Layer & Name, Position, Normal, Type, and Order \\
\bottomrule
\end{tabular}
\end{table}

\subsection{Construction Robot Motion Planning and Execution Control Module}\label{construction-robot-motion-planning-and-execution-control-module}

This module receives and leverages data from BIM to inform the construction robot's decision-making and execution processes, enabling the robot to effectively plan and carry out construction tasks without collision. Specifically, the module processes BIM data to update the collision map and generate trajectories for each assembly target in order based on the method proposed in \cite{wang2024}. Mesh models from the As-built layer are integrated as collision objects, while additional collision objects for materials are created using either detailed mesh data or standard material dimensions based on their type, along with specified position and orientation attributes. These collision objects form a detailed planning scene representation in MoveIt, allowing for collision-free robot motion planning. The construction robot's picking and placing poses are computed by matching the next target in the sequence with its material type, using BIM provided target and corresponding material position and orientation information. Based on the generated end-effector poses, the module generates the corresponding joint space trajectories for picking up the material and placing it at the target pose. These trajectories will then be published and serve as the reference for viewpoint selection.

Apart from construction robot motion planning, this module has another critical functionality: transforming the BIM data into a 3D OGM representing the environment. Points are sampled on mesh models obtained from the BIM to create a point cloud. This point cloud is then converted into a 3D OGM, denoted as \(\text{M}(E)\). In this grid mapping, cells containing points are marked as occupied, while empty cells are marked as free. This detailed spatial mapping of the environment is sent to the Viewpoint Selection Module for visibility assessment and collision check. Additionally, once supervising robots position their vision sensors at the selected viewpoints, this module will control the construction robot to execute the generated motion plan. After execution, this module will update the collision objects in the planning scene and the OGM to reflect material removal and repositioning at the target installation position. 

\subsection{Viewpoint Selection Module}\label{optimal-viewpoints-selection-module}

This module utilizes the generated next-step motion plan of the construction robot to automatically select viewpoints for supervising robots. First, potential camera poses of vision sensors mounted on supervising robots' end effectors are sampled in the supervising robots' configuration space within the work area. Initial filtering is conducted to filter unfeasible poses that lack coverage of the target area or have pose collision. Following that, a viewpoint determination methodology is proposed. In the first stage, candidate selections are formed by optimizing proximity and coverage, while the second stage prioritizes the one with the highest visibility of the target object. The technical approach is detailed in Section~\ref{optimal-view-selection}. The viewpoint selection results are then sent to the Supervising Robot Control Module for execution.

\subsection{Supervising Robot Control module}\label{supervising-robot-control-module}
The control module handles task allocation and coordinates the movement of supervising robots, which can be either mobile manipulators or drones. This section uses the mobile manipulator as a primary example, as it presents additional complexity by requiring control of both the mobile base and the arm. The module receives the selected viewpoints from the Viewpoint Selection Module and generates control commands for the mobile manipulators to position their cameras accordingly. The mobile manipulator control is divided into two parts: single mobile manipulator control and task allocation for multi-supervising robots.

\subsubsection{Single Mobile Manipulator Control}\label{single-mobile-manipulator-control}

For mobile manipulators, the control module handles both the navigation of the mobile base and the motion of the manipulator. ROS navigation stack \cite{ros_navigation_stack} is used for mobile base navigation. We construct the 2D OGM for navigation by projecting the 3D OGM generated in Section~\ref{construction-robot-motion-planning-and-execution-control-module}, which reflects the surroundings and the construction robot's current status. For global path planning, the system utilizes Navfn, a grid-based planner implementing Dijkstra's algorithm \cite{dijkstra2022note} to calculate the shortest path from the robot’s current position to its goal on the OGM. For local path planning, the system uses the Timed Elastic Band (TEB) planner, which optimizes the robot's trajectory in real-time. This online local planner allows quick path adjustments in response to the dynamic environment, enabing smooth and efficient navigation. The manipulator's motion planning is handled by the MoveIt \cite{coleman2014reducing}, which provides tools for computing kinematics and ensuring precise control of the camera's position. Together, these components ensure that both the mobile base and the manipulator work in harmony to position supervising robots' cameras at desired viewpoints.

\subsubsection{Task Allocation for Multi-Supervising Robots}\label{task-allocation-for-multi-supervising-robots}

Multiple supervising robots can be incorporated into the framework. In this paper, two supervising robots are involved, but the task allocation method can scale to include additional supervising robots. The proposed task allocation method is based on the navigation cost to each destination.

The viewpoint selection results contain candidates that are either a single viewpoint or a combination of multiple viewpoints. The task allocation problem is to assign viewpoints to the supervising robots. The system considers two scenarios: 1) only one single viewpoint is selected and 2) a combination of viewpoints is selected. When a single viewpoint is selected, one of the supervising robots is assigned to position its camera at this viewpoint. In the second scenario, when a pair of viewpoints is selected, each of the two supervising robots positions its camera at one of the viewpoints. In this scenario, efficient assignment of the two viewpoints to the two robots is crucial.

This allocation problem is indeed a Linear Assignment Problem (LAP). The LAP involves assigning a set of tasks to a set of agents to minimize the total cost, ensuring that each agent is assigned to exactly one task and each task is assigned to exactly one agent \cite{chakraa2023}. Given that only two tasks (viewpoints) need to be assigned to two supervising robots, the dimension of the cost matrix is 2×2, and the total costs for the two possible assignments are compared. However, this method can be scaled up to accommodate more involved supervising robots and viewpoints by using optimization methods like the Hungarian algorithm to solve large LAPs. To simplify the task allocation, only the mobile base’s navigation cost is considered, as the primary complexity and cost are associated with the base's movement. In this paper, the global path planning result is used to calculate the Euclidean distance of the path for each robot-viewpoint pair. These distances serve as the costs in the assignment problem. The total cost for each possible combination of assignments is then calculated, and the combination with the lowest total cost is chosen as the solution.

\section{Technical Approach for Viewpoint Selection}\label{optimal-view-selection}

The key challenge in coordinating multiple supervising robots to observe the construction robot task-performing process effectively lies in selecting viewpoints for vision sensors on supervising robots. This paper primarily focuses on on-site construction assembly tasks, specifically large-scale object installation, as it poses significant occlusion and out-of-view challenges for visual perception. Unlike the well-established art gallery framework that solves the camera placement problem by maximizing the visual coverage of a static area \cite{fleishman2000}, our approach targets enhancing visual perception of a dynamic robotic construction process in which both the robots and manipulated objects are in motion.

In this paper, the vision sensors on supervising robots capture real-time visual data, including images and point clouds, for perceiving the robotic construction process. This allows the construction robot to maintain awareness of its surroundings, its own components, and the materials it manipulates, enabling it to adapt to changing conditions. To effectively capture the robotic operation process, the proposed viewpoint selection method incorporate three key considerations: 

\begin{itemize}
    \item Coverage of the area of interest, specifically the motion envelope of the robot, which represents the spatial region that the robot and its manipulated object will occupy along the planned trajectory. This region is crucial and must be monitored for any changes, such as the presence of human workers or obstacles.

    \item Proximity to the subject, so that a sufficient level of detail can be captured. The effectiveness of the computer vision-based methodology heavily depends on the pixel resolution of the subject (i.e., area of interest or target object) in the captured images. 
    
    \item Visibility of the target object during task execution, typically referring to features of interest for specific tasks that need to be consistently tracked (e.g., the construction material being installed). Maintaining visibility of these elements is essential for monitoring the installation process and detecting possible collisions or interactions between the material and the environment.
\end{itemize}

\begin{figure}[h!]
  \centering
  \begin{tikzpicture}[
      scale = 0.3,
      node distance=1.1cm and 2.2cm,
      every node/.style={rectangle, rounded corners, minimum width=1cm, minimum height=0.8cm, align=center, font=\scriptsize\bfseries},
      process/.style={fill=softteal, text=black},
      substep/.style={fill=softgreen, text=black},
      filter/.style={fill=gray!60, text=white},
      evaluate/.style={fill=softpurple, text=black},
      rank/.style={fill=softorange, text=black},
      arrow/.style={thick,->,>=stealth}
  ]

  \node[process] (sampling) {Sampling\\candidate viewpoints};
  \node[substep, below=of sampling] (trajectory) {Evaluate\\robot's motion envelope\\coverage};
  \node[substep, below=of trajectory] (features) {Evaluate\\target object coverage};
  \node[substep, below=of features] (collision) {Check collision};

  \node[draw, dashed, fit=(trajectory) (features) (collision)] (filtering) {};
  \node[anchor=south west, font=\bfseries, xshift = 0.2cm] at (filtering.north west) {\scriptsize Initial filtering};

  \node[process, right=of sampling, yshift=1cm] (candidate) {Candidate viewpoints};
  \node[filter, below=of candidate] (optimization) {Multi-objective optimization\\ (distance \& \\ motion envelope coverage)};
  \node[process, below=of optimization] (solutions) {A subset of solutions};

  \node[evaluate, below=of solutions] (evaluate) {Evaluate visibility\\of each solution};
  \node[rank, below=of evaluate] (rank) {Rank and select};

  \draw[arrow] (sampling) -- (trajectory);
  \draw[arrow] (trajectory) -- (features);
  \draw[arrow] (features) -- (collision);

  \draw[arrow] (collision.east) -- ++(4,0) |- (candidate.west);
  \draw[arrow] (candidate) -- (optimization);
  \draw[arrow] (optimization) -- (solutions);

  \draw[arrow] (solutions) --(evaluate);
  \draw[arrow] (evaluate) -- (rank);

  \end{tikzpicture}
  \caption{Flowchart for candidate viewpoints selection process}
  \label{fig:view_selection}
\end{figure}
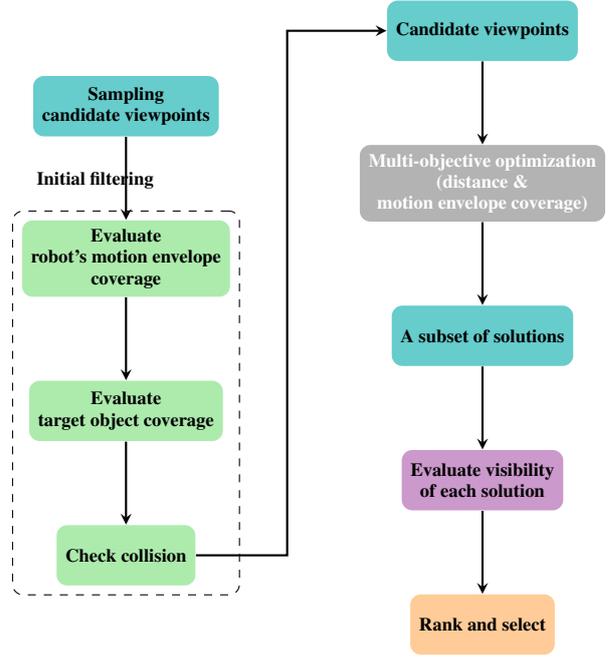

In addition, viewpoints must satisfy a collision-free constraint, allowing supervising robots to safely position themselves at the viewpoint. To achieve the objectives mentioned above, a viewpoint selection method is proposed (Figure~\ref{fig:view_selection}). Camera poses are first sampled in the supervising robots' configuration space and then go through initial filtering to ensure that the candidate viewpoints provide sufficient coverage and are reachable by the supervising robots without collision. The selected viewpoints then undergo two stages of further optimization and evaluation to determine the final outcome, consisting of either a single view or a combination of \(N\) views each assigned to a supervising robot. This paper limits $N$ to 2 for practical considerations of affordability and cost.

\subsection{Mathematical Modeling}\label{mathematical-modeling}

\subsubsection{Camera View Modeling}\label{camera-view-modeling}

\begin{figure}[h!]
  \centering
    \includegraphics[width=\dimexpr\columnwidth]{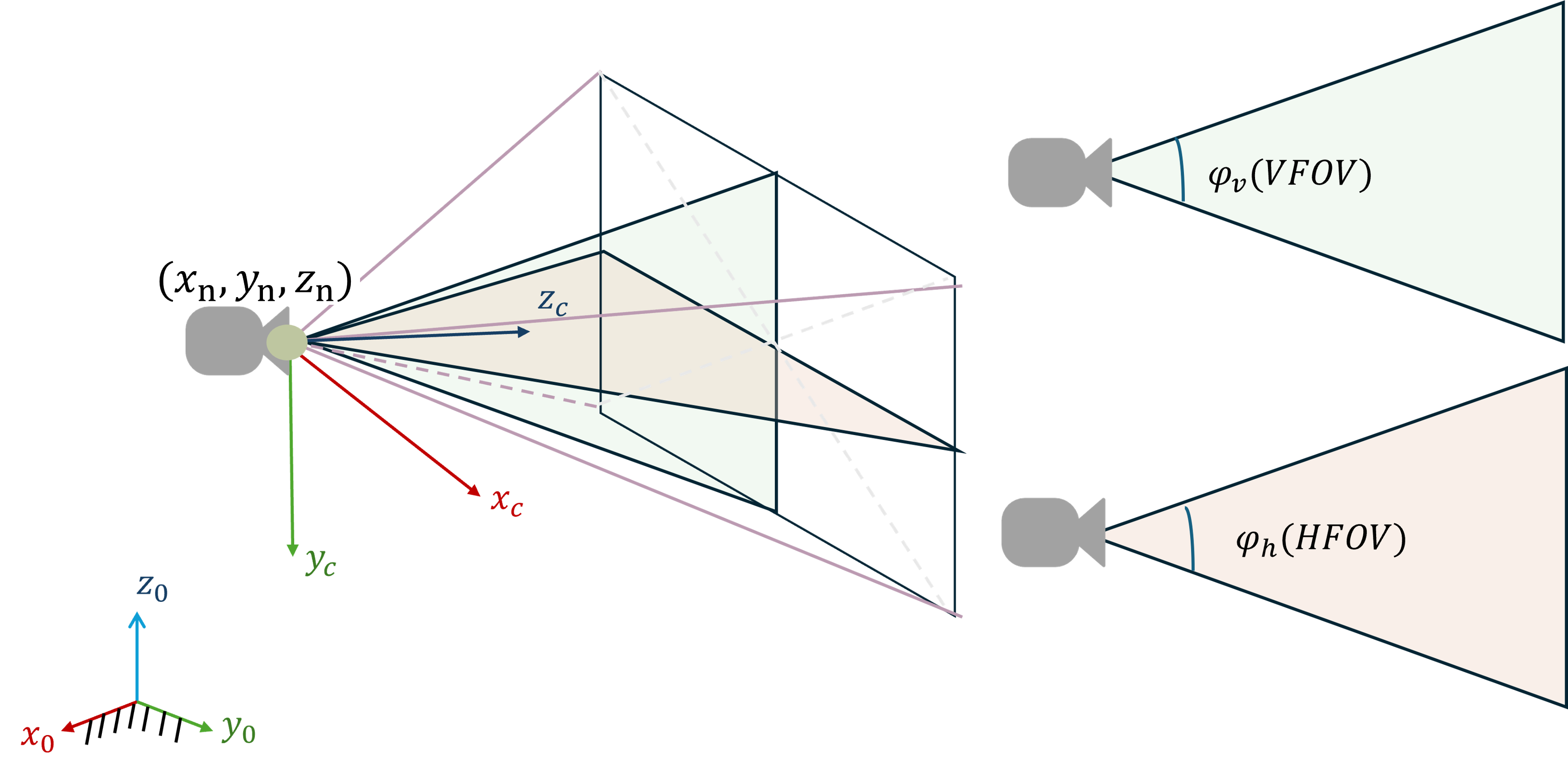}
  \caption{Illustration of camera view modeling}
  \label{fig:cam-modeling}
\end{figure}

View of Camera \(n\), \(\boldsymbol{v_{n}} \subset \mathbb{R}^{3}\), is modeled as a truncated cone with vertex at \(\left( x_{n},y_{n},z_{n} \right)\) and is defined by the vertical FOV angle \(\varphi_{v}\), and horizontal FOV angle \(\varphi_{h}\), as shown in Figure~\ref{fig:cam-modeling}. The orientation of the camera is given by  \(\left( \alpha_{n},\beta_{n},\gamma_{n} \right)\), representing the rotation around the three coordinate axes (roll, pitch, and yaw) relative to the global reference \(\left\{ 0 \right\}\ \). Therefore, the view of Camera \(n\) is characterized as \(\boldsymbol{v_{n}} = \left\{ x_{n},\ y_{n},z_{n},\alpha_{n},\beta_{n},\gamma_{n},\varphi_{v},\varphi_{h} \right\}=\{\boldsymbol{\left(vp_n\right)}, \varphi_{v},\varphi_{h}\}\) , where $\boldsymbol{\left(vp_n\right)}$ represents the campera pose corresponding to the $\boldsymbol{v_{n}}$. The combined view of \(N\) camera sensors is represented as $V = \bigcup_{n = 1}^{N} \boldsymbol{v_{n}}$, $n \in \{1, 2, \dots, N\}$.

\subsubsection{Robot Motion Envelope Representation}

Let \(G\left(\boldsymbol{\theta_{i}}\right) \subset \mathbb{R}^{3}\) be the geometric region occupied by the robot and its manipulated object in 3D Euclidean space at a specified joint state \(\boldsymbol{\theta_{i}}\). The robot's motion envelope along its planned trajectory is denoted as $G\left( \boldsymbol{s} \right)$, where \(\boldsymbol{s} = \left\{ \boldsymbol{\theta_{1}}, \boldsymbol{\theta_{2}}, \ldots, \boldsymbol{\theta_{K}} \right\}\) represents the joint space trajectory $\boldsymbol{s}$ consisting of \(K\) discrete joint states. To represent \(G\left( \boldsymbol{s} \right)\), we use point clouds derived from the geometry of robot links and the manipulated objects at various joint states along the trajectory $\boldsymbol{s}$. Given a joint state \(\boldsymbol{\theta_{i}}\in \boldsymbol{s}\), the transformation of the $j$-th link $T_{j,\ \boldsymbol{\theta_{i}}} \in SE(3)$ with respect to the global coordinate \(\left\{ 0 \right\}\ \)is determined using forward kinematics.  Here, the term ``link'' refers to either a robot link or the manipulated object. By applying the transformation \(T_{j,\ \boldsymbol{\theta_{i}}}\) to the initial geometry \(G_{j,\ init}\), the geometry of link \(j\) at joint state \(\boldsymbol{\theta_{i}}\) is obtained and denoted as \(G_{j,\ \boldsymbol{\theta_{i}}\ } = T_{j,\ \boldsymbol{\theta_{i}}}\left( G_{j,\ init} \right)\).

\begin{figure}[h!]
  \centering
  \includegraphics[width=0.8\dimexpr\columnwidth]{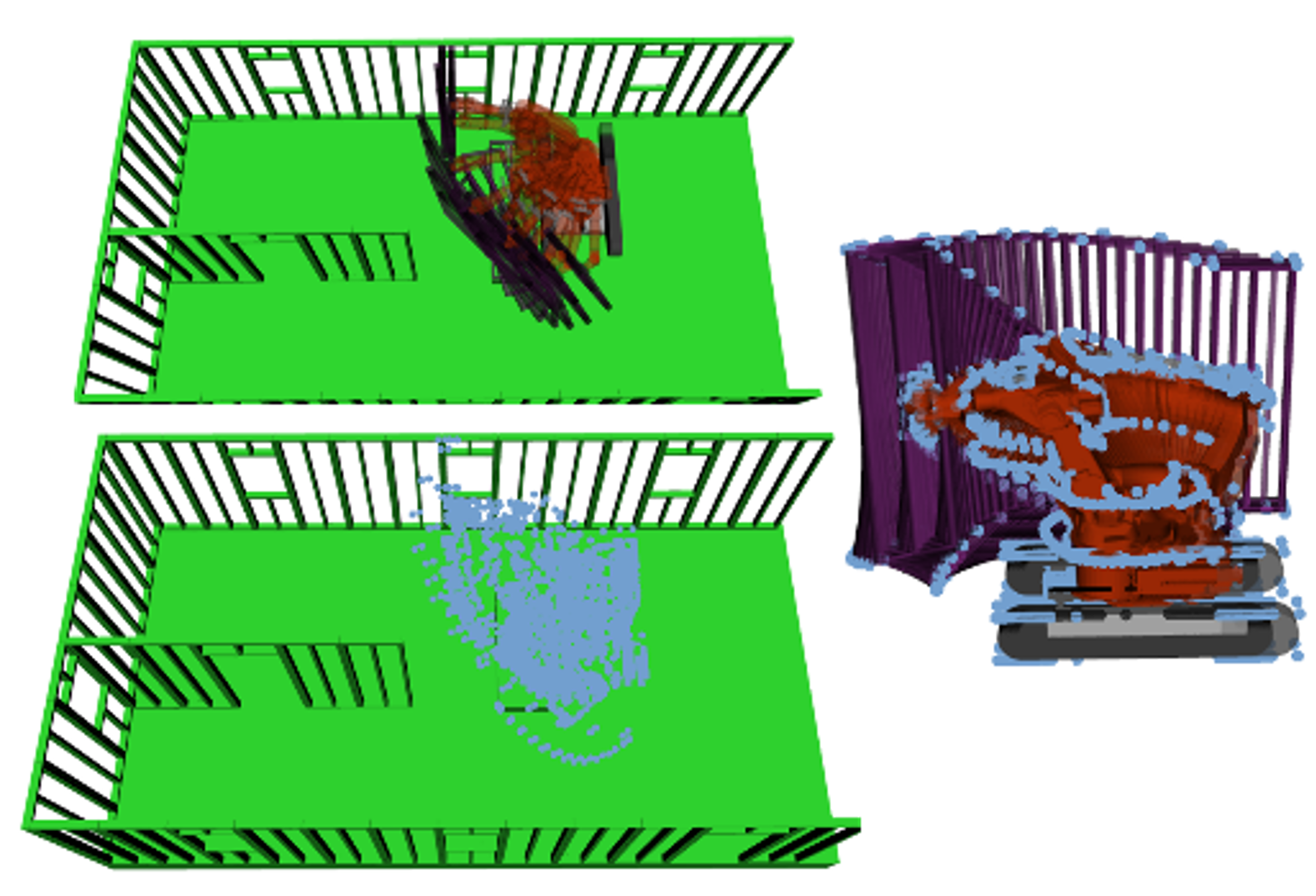}
  \caption{Illustration of robot motion envelope representation}
  \label{fig:geometry-modeling}
\end{figure}

Then, the Oriented Bounding Box (OBB) of \(G_{j,\ \boldsymbol{\theta_{i}}}\)  provides a tight-fitting rectangular boundary around the geometry, offering a simplified approximation of the space occupied by $j$-th link at $\boldsymbol{\theta_i}$. The OBB is defined by its eight corner points, which denote the extremities of the bounding volume. The corner points of \(G_{j,\ \boldsymbol{\theta_{i}}}\)are denoted as $\boldsymbol{cp_{j,\boldsymbol{\theta_{i}}}} = \{cp_{j,\boldsymbol{\theta_{i}}}^{1}, \cdots,cp_{j,\boldsymbol{\theta_{i}}}^{8}\}$. Therefore, the motion envelope \(G\left( \boldsymbol{s} \right)\) takes the union of the corner points from OBBs of all the links across all joint states along the trajectory:

\begin{equation}
  G\left( \boldsymbol{s} \right) = \bigcup_{i=1}^{K}{\bigcup_{j = 1}^{L}\boldsymbol{cp_{j,\boldsymbol{\theta_{i}}}}}
  \label{eq:G(s)}
\end{equation}
where $L$ refers to the total number of links, including the manipulated object. An example of this representation is illustrated as the blue point cloud in Figure~\ref{fig:geometry-modeling}.

\subsubsection{Coverage Modeling}\label{coverage-modeling}

Let $V_m\in  \mathbf{V}$ denote one possible viewpoint combination. The coverage of $V_m$ is defined as:

\begin{equation}
  C(V_m) = \frac{\mu\left( G\left(\boldsymbol{s} \right) \cap V_m \right)}
       {\mu\left( G\left( \boldsymbol{s} \right) \right)}
  \label{eq:C}
\end{equation}

where \(\mu\left( \cdot \right)\) is the measurement function.

The total measure \(\mu\left( G\left( \boldsymbol{s} \right) \right)\) can be approximated by the number of points in \(G\left( \boldsymbol{s}\right):\ \mu(G\left( \boldsymbol{s} \right)) = |G\left( \boldsymbol{s} \right)| = 8KL\). The measure \(\mu(G(\boldsymbol{s}) \cap V_m)\) is given by:
\[
\mu(G(\boldsymbol{s}) \cap V_m) = \sum_{\mathbf{p} \in G(\boldsymbol{s})} C_{V_m}(\mathbf{p}),
\]
where the indicator function \(C_{V_m}(\mathbf{p})\) is defined as:
\[
C_{V_m}(\mathbf{p}) =
\begin{cases} 
1, & \text{if } \mathbf{p} \in V_m, \\
0, & \text{otherwise}.
\end{cases}
\]

Therefore, the calculation of coverage in Eq.~\ref{eq:C} can be further expressed as:

\begin{equation}
  C{(V_m)} =
  \frac{\sum_{\mathbf{p} \in G(\boldsymbol{s})} C_{V_m}(\mathbf{p})}
       {8KL}
  \label{eq:CVm}
\end{equation}

\subsubsection{Visibility Modeling}\label{visibility-modeling}

Let \(G_{obj,\ \boldsymbol{\theta_{i}}\ }\) represent the geometry of the target object at \(\boldsymbol{\theta_{i}}\), and \(\text{Vis}\left( G_{obj,\ \boldsymbol{\theta_{i}}\ }, V_m \right)\) denote the visibility metric that quantifies the extent to which \(G_{obj,\ \boldsymbol{\theta_{i}}\ }\)is visible from the $V_m$ at \(\boldsymbol{\theta_{i}}\). \(\text{AvgVis}\left( V_m \right)\) is the average visibility of the target object over the whole trajectory \(\boldsymbol{s}\) under combined view \(V_m\).

\begin{equation}
  \text{AvgVis}\left(V_m \right) = \frac{1}{K}\sum_{i = 1}^{K}{\text{Vis}\left( G_{obj,\ \boldsymbol{\theta_{i}}}, V_m\right)}
  \label{eq:AvgVis}
\end{equation}
where \(K\) is the number of joint states along the trajectory. The ray casting method is used to evaluate \(\text{Vis}\left( G_{obj,\ \boldsymbol{\theta_{i}}} , V_m\right)\), as detailed below.

\vspace{0.5\baselineskip}
\noindent(1) Spatial representation for visibility modeling

To evaluate the visibility of the \(G_{obj,\ \boldsymbol{\theta_{i}}}\), we construct 3D OGM \(\text{M}\left(\boldsymbol{\theta_{i}}\right)\) that maps all the related entities at $\boldsymbol{\theta_{i}}$, including the environment, robot, and the manipulated object. Each voxel in $M\left(\boldsymbol{\theta_{i}}\right)$ is defined as:

\begin{equation}
  M\left(\boldsymbol{\theta_{i}}\right) = 
  \begin{cases} 
    1, & \text{if } M(E) = 1 \text{ or } \\
       & \quad \text{voxel lies on surface of any } \text{OBB}\left( G_{j,\ \boldsymbol{\theta_{i}}}\right),\\ 
       & \quad j = 1,\dots,m \\ 
    0, & \text{otherwise} 
  \end{cases} 
  \label{eq:eq5}
\end{equation}

\begin{figure*}[h!]
  \centering
  \includegraphics[width=0.9\textwidth]{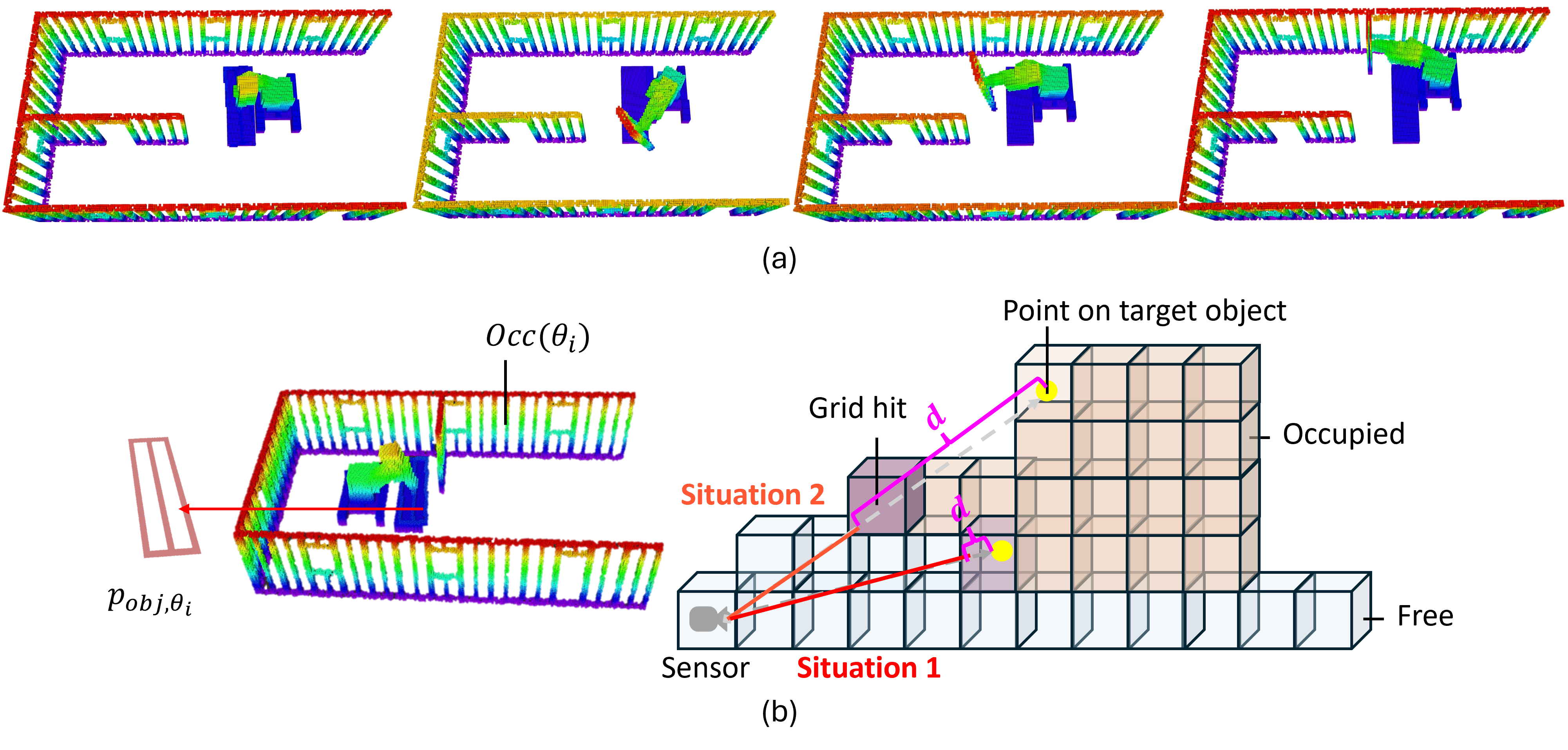}
  \caption{Illustration of generated OGM and ray-casting-based visibility. (a) OGM representation of related entities with different robot joint states. (b) Visibility status determination by ray-casting from the sensor to the measured point.}
  \label{fig:occupancy_visibility}
\end{figure*}

Here, $M(E)$ represents the environmental OGM constructed in Section~\ref{construction-robot-motion-planning-and-execution-control-module}. Figure~\ref{fig:occupancy_visibility}a illustrates the example OGM at several joint states.

\vspace{0.5\baselineskip}
\noindent(2) Target points sampling

The OGM alone does not provide sufficient information to distinguish the exact voxels belonging to the target object, posing a challenge for ray-casting-based visibility checks. To address this, a 3D point set \(P_{obj,\ \boldsymbol{\theta_{i}}}\) representing points on the target object at each \(\boldsymbol{\theta_{i}}\) is recorded. To ensure the set of points remain consistent across all joint states for visibility evaluation, the points are first uniformly sampled from the surface of the object's geometry in its local frame and then transformed based on the calculated transformation relative to the global frame \(\left\{ 0 \right\}\).

\vspace{0.5\baselineskip}
\noindent(3) Ray-casting based visibility evaluation

A ray-casting method is employed to assess the visibility of each point \(p_{k} \in P_{obj,\ \boldsymbol{\theta_{i}}}\) (Figure~\ref{fig:occupancy_visibility}b). For each view \(\boldsymbol{v_{n}} \in V_m\), each point \(p_{k}\ \)is first checked to determine whether it lies within the view \(\boldsymbol{v_{n}}\). If \(p_{i}\ \text{within }\boldsymbol{v_{j}}\), ray casting from the vertex point \(\left( x_{n},\ y_{n},z_{n} \right)\ \)of \(\boldsymbol{v_{n}}\) to \(p_{k}\) is used to determine whether any objects or obstacles obstruct the ray's path. First, the Euclidean distance \(d\left( p_{k}, \text{voxel}_{\text{hit}} \right)\ \) between the target point \(p_{k}\) and the center of the terminal grid cell \(\text{voxel}_{\text{hit}}\) (i.e., the point where the ray intersects an occupied grid in \(M\left(\boldsymbol{\theta_{i}}\right)\ \)) is calculated. The visibility \(\text{Vis}\left( p_{k},\ \boldsymbol{v_{n}} \right)\) of \(p_{k}\) from \(\boldsymbol{v_{n}}\ \)is then determined by comparing the distance \(d\left( p_{k},\text{voxel}_{\text{hit}} \right)\) to a predefined distance threshold \(\epsilon\):

\begin{equation}
  \text{Vis}\left( p_{k}, \boldsymbol{v_{n}} \right) = 
  \begin{cases} 
    1, & \text{if } d\left( p_{k}, \text{voxel}_{\text{hit}} \right) \leq \epsilon \text{ and } p_{k}\ \text{in}\  \boldsymbol{v_{n}} \\ 
    0, & \text{otherwise}
  \end{cases}
  \label{eq:eq6}
\end{equation}

If \(d\left( p_{k},\text{voxel}_{\text{hit}} \right)\) is sufficiently small (Situation 1 in Figure~\ref{fig:occupancy_visibility}b),  $p_k$ is considered visible under \(v_{n}\). Otherwise, the point is considered occluded (Situation 2 in Figure~\ref{fig:occupancy_visibility}b). The overall visibility \(\text{Vis}\left( G_{obj,\ \boldsymbol{\theta_{i}}\ }, V_m \right)\ \)of a target given $V_m$ at the joint state \(\boldsymbol{\theta_{i}}\) is calculated as the proportion of points in the set \(P_{obj,\ \boldsymbol{\theta_{i}}}\text{\ \ }\) that are visible under any camera views \(\boldsymbol{v_n} \in V_m\):

\begin{equation}
  \text{Vis}\left( G_{obj,\ \boldsymbol{\theta_{i}}\ }, V_m \right) = \ \frac{1}{\left| P_{obj,\ \boldsymbol{\theta_{i}}} \right|}\sum_{p_{k} \in P_{obj,\ \boldsymbol{\theta_{i}}}}^{}{\underset{\boldsymbol{v_{n}} \in V_m}{\text{ma}x}{\text{Vis}\left( p_{k},\ \boldsymbol{v_{n}} \right)}}
  \label{eq:eq7}
\end{equation}

where \(\underset{\boldsymbol{v_{n}} \in V_m}{\text{ma}x}{\text{Vis}\left( p_{k},\ \boldsymbol{v_{n}}\right)}\) captures the maximum visibility value for each point \(p_{k}\ \)across all views \(v_{n}\) in $V_m$. This accounts for the point being visible in any of the views $\boldsymbol{v_{n}}$. \(\left| P_{obj,\ \boldsymbol{\theta_{i}}} \right|\) is the total number of points in the target point set \({P_{obj,\ \boldsymbol{\theta_{i}}}}
\).

\subsubsection{Distance Modeling}

To increase the pixel resolution of the subject in the capture, we aim to minimize the distance between the camera and the target. Specifically, in this paper, we focus on the distance between the camera and the robot envelope for the picking operation and the distance between the camera and the manipulated object along its trajectory during the placing operation. 

For the picking operation, we define the distance as the average Euclidean distance between \(Centriod\left( G\left( \boldsymbol{s} \right)\right) \in \mathbb{R}^3\) and the vertex \((x_{n},y_{n},z_{n})\) of each camera view \(\boldsymbol{v_{n}}\):

\begin{equation}
  \text{d}\left( G(\boldsymbol{s}),V_m \right) = \frac{1}{N}\sum_{v_{n} \in V}^{}{d(Centriod(G\left(\boldsymbol{s}\right),\ (x_{n},y_{n},z_{n}))}
  \label{eq:eq8}
\end{equation}
For the placing operation, the point set \(P_{obj,\ \boldsymbol{\theta_{i}}}\) that represents the manipulated object at \(\boldsymbol{\theta_{i}}\) is obtained in \ref{visibility-modeling}. The distance is defined as the average Euclidean distance between each \(Centroid(P_{obj,\ \boldsymbol{\theta_{i}}})\) across all joint states in  $\boldsymbol{s}$ and all camera positions in $V_m$.

\begin{equation}
 d\left(\text{obj}, V_m \right) = \frac{1}{N} \sum_{\boldsymbol{v_{n}} \in V} \frac{1}{K} \sum_{\boldsymbol{\theta_{i}} \in \boldsymbol{s}} d\left( Centroid\left( P_{\text{obj}, \boldsymbol{\theta_{i}}} \right), (x_{n}, y_{n}, z_{n}) \right)
  \label{eq:eq9}
\end{equation}

\subsection{Camera Pose Sampling}\label{camera-pose-sampling}

Both mobile manipulators and drones could serve as supervising robots. In this paper, we will use mobile manipulators as an example. For the mobile manipulator, while the mobile base effectively eliminates the workspace limitations of the manipulator by providing greater mobility, determining a target pose and solving for the robot's joint states using inverse kinematics can be computationally expensive, especially when the robotic arm has limited degrees of freedom. Therefore, it is crucial to consider supervising robot's configuration when sampling the possible camera poses. Additionally, ensuring a well-distributed scatter of the robot's mobile base positions is necessary to achieve diverse views within the work area boundaries. To solve these challenges, the proposed method introduces a configuration-based camera pose sampling approach and a filtering criterion to pre-filter the camera poses before selection.  Camera pose sampling begins by sampling the mobile base positions within the operational area boundary and the arm joint states within their limits. Forward kinematics is then used to determine the corresponding camera poses based on the sampled position and joint states. A pre-filtering criterion excludes camera poses that do not face the anticipated robot motion envelope.

\subsubsection {Configuration-based Camera Poses Sampling}

The arm's configuration space sampling is relatively straightforward, with each joint value being randomly sampled within its joint limits. However, sampling the base position is more complex due to the need to ensure the base stays within the often irregular operational area boundary. To address this, the process utilizes BIM to extract a detailed floor mesh. This floor mesh is processed to define the boundary for base coordinate sampling. Using this boundary, valid base positions are sampled to ensure they lie within the defined area.

\begin{figure}[h!]
  \centering
  \includegraphics[width=\dimexpr\columnwidth]{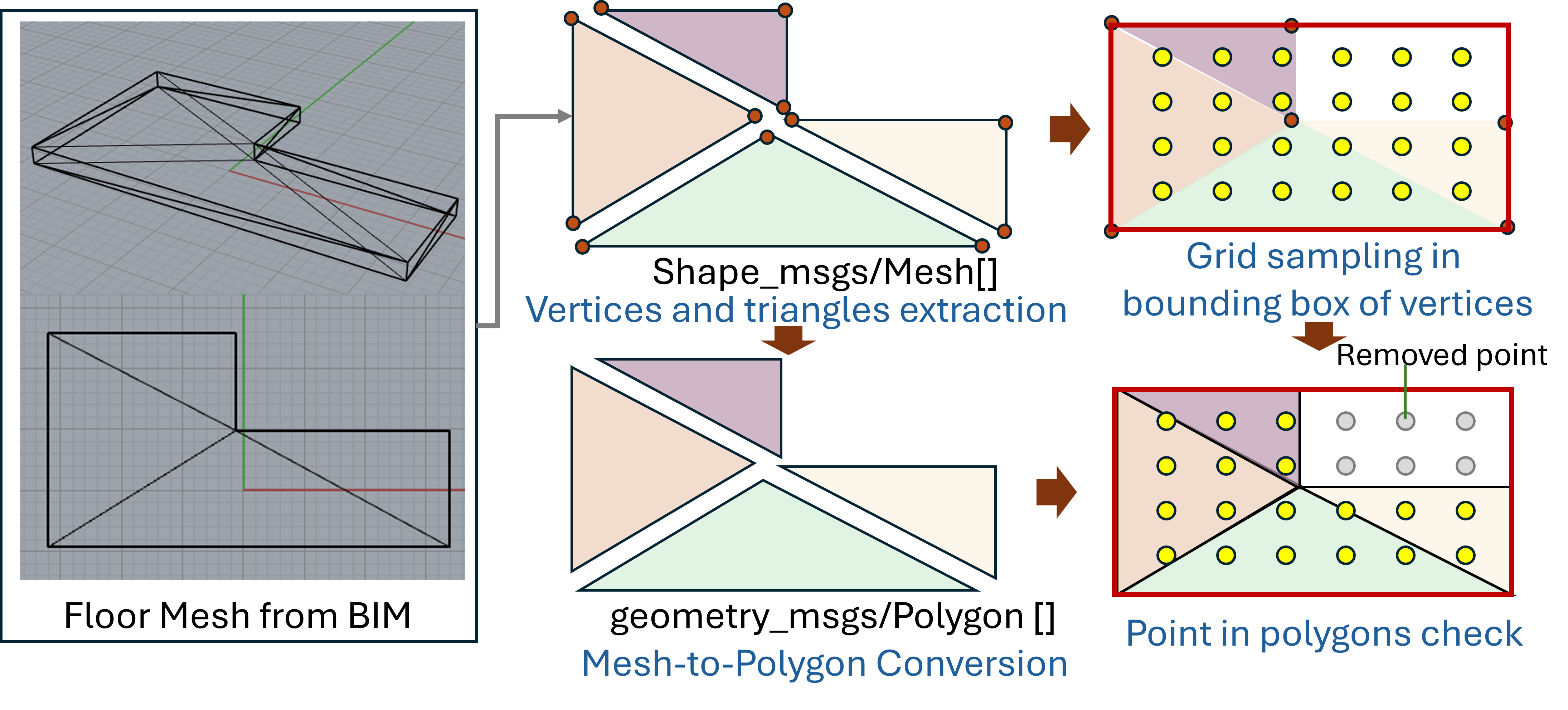}
  \caption{Supervising robot base coordinates sampling with BIM floor mesh}
  \label{fig:base-sampling}
\end{figure}

As illustrated in Figure~\ref{fig:base-sampling}, the floor mesh is first processed to extract the vertices and triangles at the height of its top surface, isolating the relevant 2D cross-section of the 3D shape. These triangles and corresponding vertices are then used to define polygons representing the sampling boundary. Once the polygons are established, a bounding box containing all the vertices of polygons is calculated. A grid sampling method is employed within this bounding box, sampling a grid of evenly spaced points. A point-in-polygon (PIP) check is performed using a ray-casting algorithm \cite{shimrat1962} to filter and record grid points within the defined polygons, which are considered as valid base position samples.

The base positions are randomly selected from these valid base position samples, with the orientation being randomly sampled within 360 degrees. Based on the combined position and orientation of the base, along with the joint states of the arm sampled within joint limits, the joint state of the supervising robot $s\theta_{n}$ is determined. Using forward kinematics, the camera's pose $\boldsymbol{\left(vp_n\right)}$ is then computed. Compared to inverse kinematics, this sampling process is more computational effective. Additionally, the sampling process ensures that the base positions are confined within the accessible operational area.

\subsubsection {Camera Pose Pre-filtering}

After the initial sampling of camera poses, a pre-filtering criterion is applied to exclude camera poses that face away from the anticipated robot's motion envelope. Specifically, a camera pose is excluded if the angle \(\alpha\) between the camera's forward-facing direction (its local z-axis in the global frame, denoted as \(\mathbf{z}_n^\text{G}\)) and the vector $\boldsymbol{r_n}$, pointing from the camera position \((x_{n}, y_{n}, z_{n})\) to the construction robot's motion envelope center \(C(G(\boldsymbol{s}))\), exceeds a specified threshold \(\alpha_{\text{threshold}}\). Here, \(\mathbf{z}_n^\text{G} = R(\alpha_{n}, \beta_{n}, \gamma_{n}) \mathbf{z}_{cn}\), where \(\mathbf{z}_{cn}\) is the z-axis of the camera in its own local frame. The rotation matrix \(R(\alpha_{n}, \beta_{n}, \gamma_{n})\) defines the camera’s orientation relative to the global reference frame. The vector $r_n$ is defined as \(\boldsymbol{r_n} = Centroid\left( G\left(\boldsymbol{s}\right) \right) - (x_{n}, y_{n}, z_{n})\). The angle \(\alpha\) between \(\mathbf{z}_n^\text{G}\) and \(r_{n}\) is then calculated as $\alpha = \tan^{-1} \left( \frac{\|\mathbf{z}_n^\text{G} \times\boldsymbol{r_n}\|}{\mathbf{z}_n^\text{G} \cdot\boldsymbol{r_n}} \right)$. The exclusion criterion is defined as:

\begin{equation}
  \text{excludePose}\left( \boldsymbol{\left(vp_n\right)}, \alpha_{\text{threshold}} \right) = 
  \begin{cases} 
      \text{true} & \text{if } \alpha > \alpha_{\text{threshold}} \\ 
      \text{false} & \text{if } \alpha \leq \alpha_{\text{threshold}}
  \end{cases}
  \label{eq:exp1}
\end{equation}

\subsection{Initial Filtering}\label{initial-filtering}

The goal of initial filtering is to ensure that the views from the sampled camera poses meet the minimum coverage requirement and satisfy the collision-free constraint. To achieve this, spatial coverage filtering and collision checking are used to filter the candidate camera poses that could be used for optimization (Figure~\ref{fig:view_selection}).

\subsubsection{Spacial Coverage Filtering}

To ensure adequate spatial coverage of the robot's motion envelope, each sampled camera pose is evaluated based on its coverage. For each view \(\boldsymbol{v_{n}}\)  corresponding to the sampled camera pose \(cp_{n}\), Eq.~\ref{eq:CVm} is used to calculate its coverage \(C(\boldsymbol{v_{n}})\). A lower bound threshold \(C_{\text{single}}\) is applied, and camera poses with coverage below this threshold are excluded:

\begin{equation}
  \text{excludePose}\left(\boldsymbol{v_{n}}, C_{\text{single}} \right) = 
  \begin{cases} 
      \text{true} & \text{if } C(\boldsymbol{v_{n}}) < C_{\text{single}} \\ 
      \text{false} & \text{if } C(\boldsymbol{v_{n}}) \geq C_{\text{single}}
  \end{cases}
  \label{eq:eq10}
\end{equation}

In addition to meeting the coverage requirement of the whole robot's motion envelope, the camera must also provide adequate coverage of the area occupied by the target object throughout the robot's operation. Following the similar approach in Eq.~\ref{eq:G(s)}, the motion envelope for the target object along the trajectory is recorded as \(G\left( \boldsymbol{s},obj \right) = \bigcup_{i=1}^{K}\boldsymbol{cp_{obj,\boldsymbol{\theta_{i}}}}\). Then, similar to the approach in Eq.~\ref{eq:CVm}, the coverage of the target object \(C_{obj}(\boldsymbol{v_{n}})\) for view \(\boldsymbol{v_{n}}\) is evaluated. Camera poses with target object coverage below a predefined lower bound $C_{\text{single, target}}$ are excluded from further consideration.

\begin{equation}
  \begin{split}
    \text{excludePose}\left(C_{\text{obj}}(\boldsymbol{v_n}), C_{\text{single, target}} \right) \\
    = 
    \begin{cases} 
        \text{true}, & \text{if } C_{\text{obj}}(\boldsymbol{v_n}) < C_{\text{single, target}}, \\ 
        \text{false}, & \text{if } C_{\text{obj}}(\boldsymbol{v_n}) \geq C_{\text{single, target}}.
    \end{cases}
  \end{split}
  \label{eq:eq11}
\end{equation}

\subsubsection{Collision Checking}

After ensuring adequate coverage, collision checking is performed to exclude sampled camera poses that result in collision. For each sampled camera pose, the robot's configuration boundary is represented by a bounding box, which is defined as an area covering a specified radius and height around the corresponding mobile base position. The radius and height are determined by the mobile base dimension and reach of the manipulator. If any part of the environment or the primary construction robot intersects with the bounding box, the camera pose is flagged as being in collision.

The process begins by checking collision with the environment. The environment's OGM \(M(E)\) is checked to determine if any grid cells fall within the defined bounding box are occupied. If any such grid cell is marked as occupied, the corresponding camera pose is flagged as in collision. 

The camera poses that pass this check are free from environmental collision and are further evaluated for potential collisions with the construction robot. First, the bounding box of the supervising robot is checked against the bounding box of the construction robot’s motion envelope \(G(\boldsymbol{s})\) for any overlap. If no overlap is detected, the camera pose is considered collision-free. If overlap exists, an additional check is conducted, where 3D OGM \(\text{M}(\boldsymbol{\theta_{i}})\) at each $\boldsymbol{\theta_{i}}$ is queried. If any grid within the bounding box of the supervising robot is marked as occupied, the camera pose is flagged as being in collision. Otherwise, it is considered collision-free. In scenarios where no target objects are defined for visibility checks such that the OGM \(\text{M}(\boldsymbol{\theta_{i}})\) is not obtained, only camera poses without bounding box overlap in the first stage are retained.

\subsection {Viewpoint Determination}\label{combination-of-camera-poses}

After the initial filtering stage, where unsuitable camera poses were eliminated, the focus now shifts to finding the single or combination of camera viewpoint(s) that maximize the coverage and proximity, and, if applicable, with maximal visibility of the target object. First, spatial coverage and distance are optimized simultaneously to generate a Pareto-optimal set of solutions, offering a balanced trade-off between these two objectives. After this, if the operation involves a target object, the solution with the highest visibility to the target object is selected from the Pareto-optimal set. Otherwise, the one with the highest coverage is selected.

\subsubsection{Maximizing FOV Coverage and Proximity Using Multi-objective Optimization}

At this stage, the goal is to find solutions that simultaneously maximize coverage and proximity (i.e., minimize distance), two crucial yet conflicting objectives. The Non-dominated Sorting Genetic Algorithm II (NSGA-II) \cite{deb2002} is employed for this purpose. NSGA-II is a popular multi-objective optimization algorithm that efficiently finds a set of Pareto-optimal solutions by balancing the trade-offs between conflicting objectives. 

\begin{figure*}[h!]
  \centering
  \includegraphics[width=0.95\textwidth]{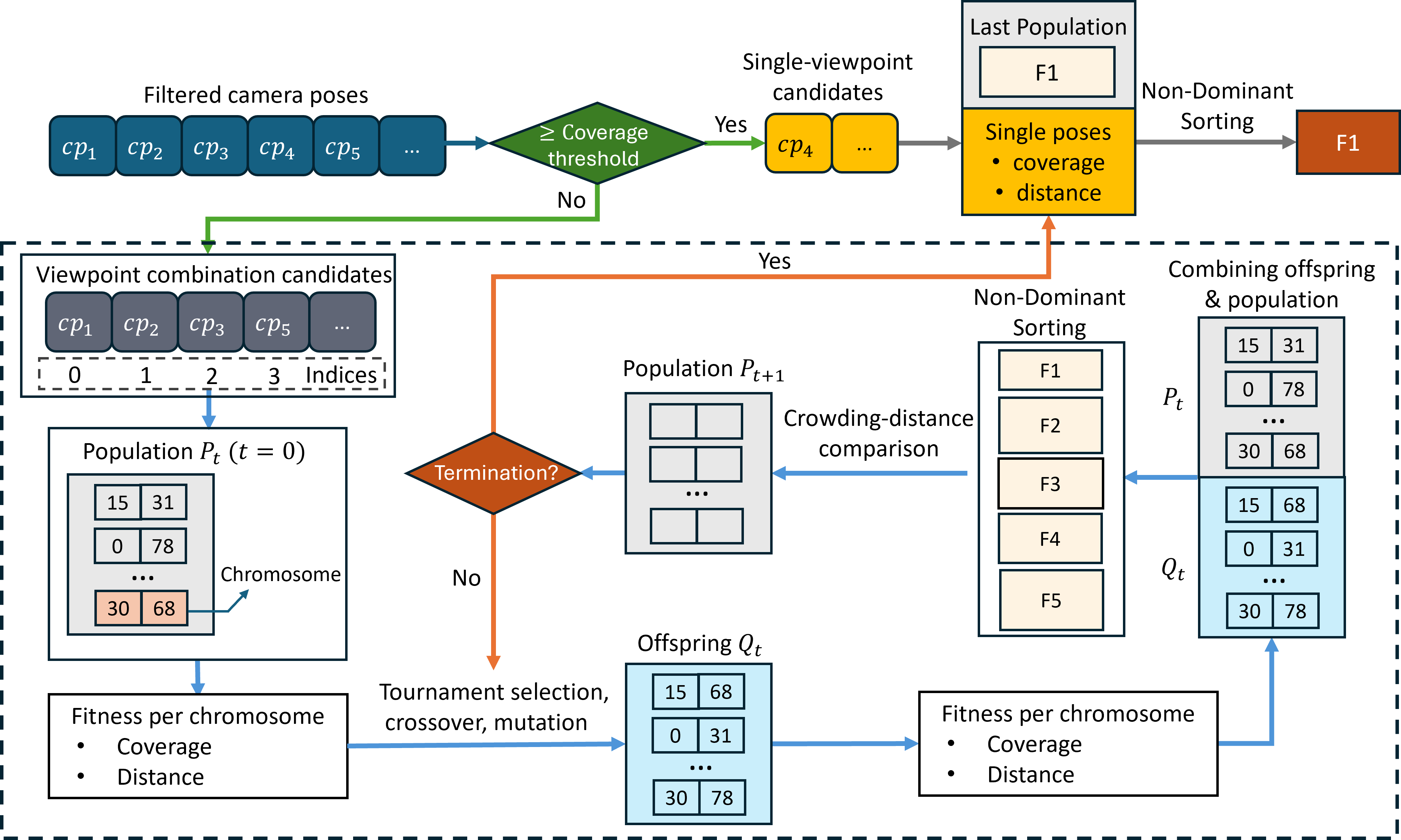}
  \caption{NSGA-II optimization process}
  \label{fig:nsga-ii}
\end{figure*}

As illustrated in Figure~\ref{fig:nsga-ii}, each camera pose \(\boldsymbol{\left(vp_{n}\right)}\) is evaluated based on its coverage \(C\left(\boldsymbol{v_n} \right)\) (Eq.~\ref{eq:CVm}). Camera poses that meet or exceed the final coverage threshold are retained as single-viewpoint candidates. These poses already satisfy the required coverage and do not need further optimization. For the camera poses that do not meet the coverage threshold individually, further optimization for finding combinations that, together, can achieve the desired coverage is required. 

The NSGA-II algorithm is employed for this combination, starting with the initial generation of a population of solutions. Each solution represents a possible combination of camera poses and is encoded as a chromosome containing camera indices from the filtered camera poses. The initial population $P_0$ consists of $N_p$ randomly generated individuals (combinations), with each chromosome representing a combination of \(N\) camera indices corresponding to the \(N\) supervising robots. In this paper, \(N =2\) since two supervising robots are considered. 

The fitness functions for optimization include spatial coverage (Eq.~\ref{eq:CVm}) and the distance to the target (Eq.~\ref{eq:eq9}) or robot's motion envelope center if no target object is specified (Eq.~\ref{eq:eq8}). The evolutionary process applies genetic operations of tournament selection, crossover, and mutation to generate new combinations (i.e., offspring $Q_t$), which are evaluated with the fitness functions. $Q_t$ are then combined with the current population $P_t$, and a fast non-dominated sorting and crowding-distance comparison are performed to select the next generation $P_{t+1}$. The process continues iteratively until the termination criteria, such as a predefined number of generation are met.

At the conclusion of the final generation, the obtained Pareto-optimal solutions are considered together with the previously retained single-view candidates and ranked again using non-dominant sorting to identify the final Pareto-optimal solutions. Finally, these solutions, representing the best possible viewpoint selection result in terms of both coverage and proximity, are further filtered by the coverage threshold to form the final candidate solution set.

\subsubsection{Maximizing Observability of the target feature}\label{max-observability}

If a target object is involved, such as the manipulated object in the placing operation, the solution with the highest visibility of the target object (quantified by Eq.~\ref{eq:AvgVis}) in the Pareto-optimal solution sets will be selected as the final solution. This approach ensures that the chosen solution is not only optimized for coverage and proximity but also considers visibility of the target object.  If no target object is involved, priority will be given to the solution with the highest coverage.

\section{Case Study}\label{case-study}

To demonstrate the feasibility of the proposed multi-robot coordination framework, a simulated construction assembly task is conducted as a proof-of-concept implementation. Specifically, prefabricated frame installation is selected as the case study as it represents a quasi-repetitive construction task that involves handling large elements and installing each element to a different location. In this case study, the KUKA KR60-3 robot mounted on a tracked mobile base is employed as the construction robot, and a TurtleBot 3 Waffle Pi equipped with an OpenManipulator-X is selected as the supervising robot. The Intel RealSense D435 camera is selected as the vision sensor. As an active stereo depth camera, it provides 3D distance measurements and RGB data. Its infrared projector enhances lightning adaptability, while its wide field of view and 10-meter maximum range make it ideal for construction tasks. Detailed specifications are available in Table~\ref{tab:realsense_specs}.

\begin{table}[h!]
  \small
  \centering
  \caption{Specifications of Intel RealSense D435}
  \label{tab:realsense_specs}
  \begin{tabular}{l l}
    \toprule
    \textbf{Features} & \textbf{Intel RealSense D435} \\
    \midrule
    \textbf{Use Environment} & Indoor/Outdoor \\
    \textbf{Maximum Depth Range} & 10 m \\
    \textbf{Depth FOV} & 87° × 58° \\
    \textbf{Depth Output Resolution} & Up to 1280 × 720 \\
    \textbf{RGB Sensor FOV} & 69° × 42° \\
    \textbf{RGB Frame Resolution} & 1920 × 1080 \\
    \bottomrule
    \end{tabular}
\end{table}

\begin{figure}[h!]
  \centering
  \includegraphics[width=0.8\dimexpr\columnwidth]{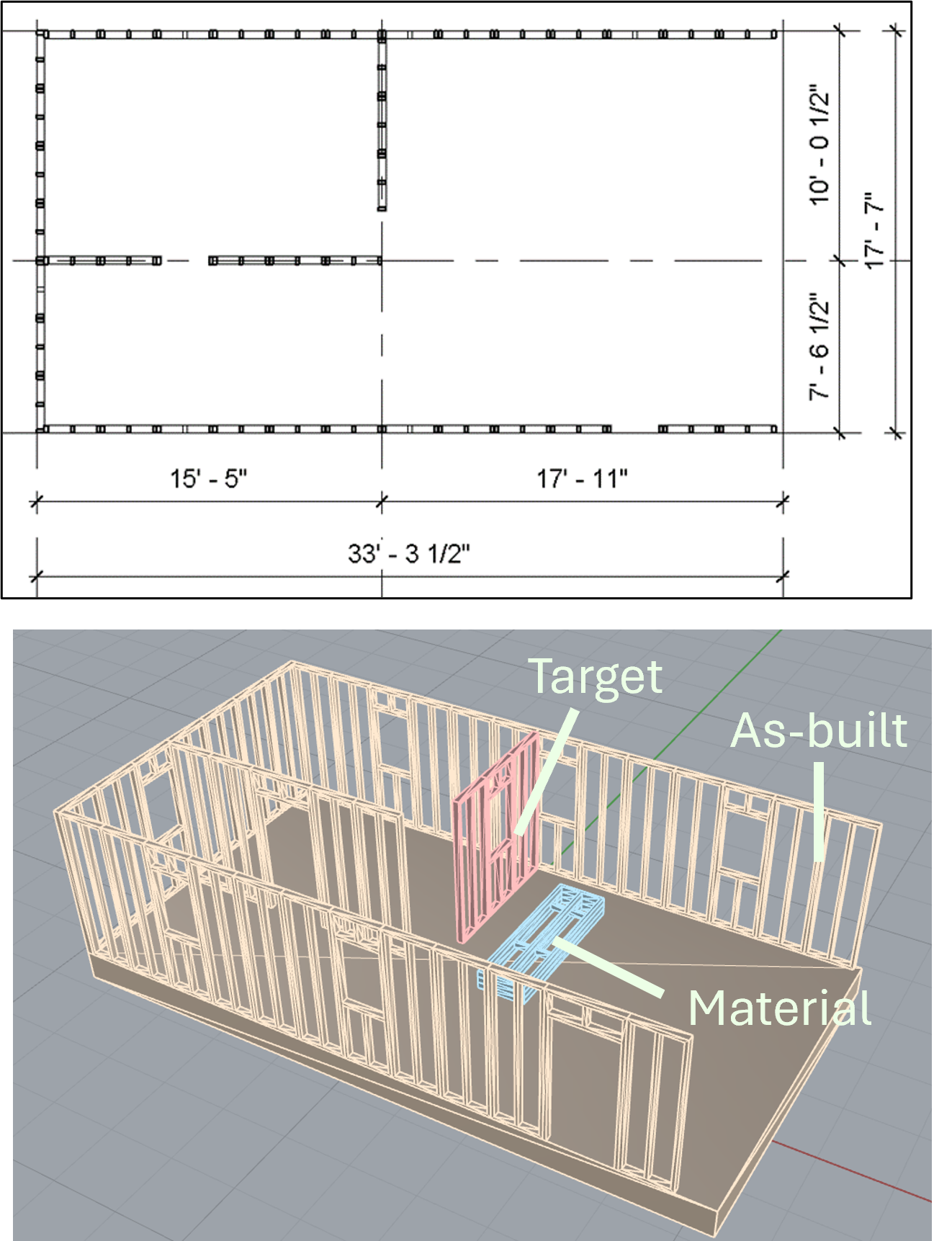}
  \caption{BIM for the Case Study}
  \label{fig:case-study-bim}
\end{figure}

In this case study, three prefabricated frames are designed to be installed, as the BIM shown in Figure \ref{fig:case-study-bim}. The different colors indicate the objects' layers in the BIM: red represents the target layer, blue indicates the material layer, and yellow indicates the as-built layer. The construction robot is supposed to install the target frames one by one based on the BIM data. For each target frame, the Construction Robot Motion Planning and Execution Control Module generates the trajectories for picking up the corresponding material and placing it to the target pose. The trajectories are then analyzed separately by the Viewpoint Selection Module to determine viewpoints for observing the construction robot's motion along the trajectory. Finally, the Supervising Robot Control Module deploys the supervising robot to the selected viewpoints. Once the supervising robots are positioned, the construction robot initiates its movement following the planned trajectory. This coordinated sequence repeats for each frame installation, cycling through the picking and placing operations. The video demonstration of the case study is available at \url{https://youtu.be/q2ZdebajGeQ}.

Two supervising robots are employed to perceive the robotic construction process. Viewpoints for supervising robots’ visual sensors are decided by the Viewpoint Selection Module. During the picking process, the supervising robots’ cameras are positioned to optimize their proximity and coverage of the construction robot's motion envelope. However, during the placing process, in addition to proximity and coverage, the visibility of the target object (i.e., the wooden frame being manipulated) is also evaluated. With the supervising robots adapting their poses to observe the picking and placing operations, high-quality visual perception is achieved throughout the entire construction process. For each motion plan, 800 camera poses are sampled in the supervising robot’s configuration space. Each camera pose is filtered by the threshold $C_{single} = 0.5$ and $C_{single,target}=0.4$, as detailed in \ref{initial-filtering}. The parameters for NSGA-II optimization (Section~\ref{combination-of-camera-poses}) are as follows: the number of generations is 70 with a population of 200 individuals, and the coverage threshold is 0.97. 



For the picking trajectory, only the proximity and coverage objectives are considered since no target object is present. Therefore, only the NSGA-II based coverage and proximity optimization is applied, with the Pareto-optimal solution achieving the highest coverage chosen. For the placing trajectory, proximity and coverage are first optimized, followed by the visibility assessment.



\begin{figure*}[h!]
  \centering
  \includegraphics[width=0.7\textwidth]{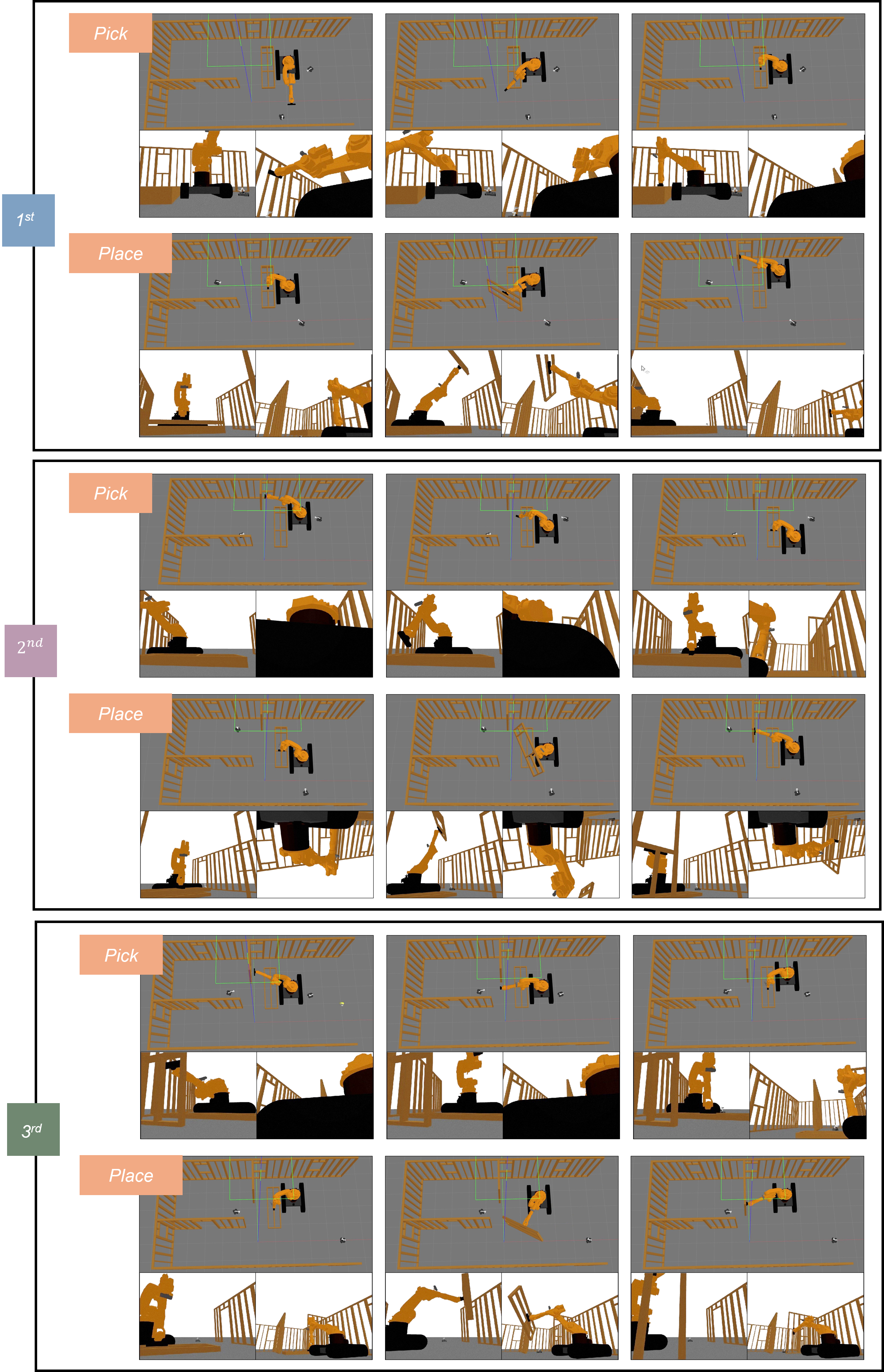}
  \caption{Real image captures from cameras on supervising robot}
  \label{fig:case-study-images}
\end{figure*}

\begin{figure*}[h!]
  \centering
  \includegraphics[width=0.7\textwidth]{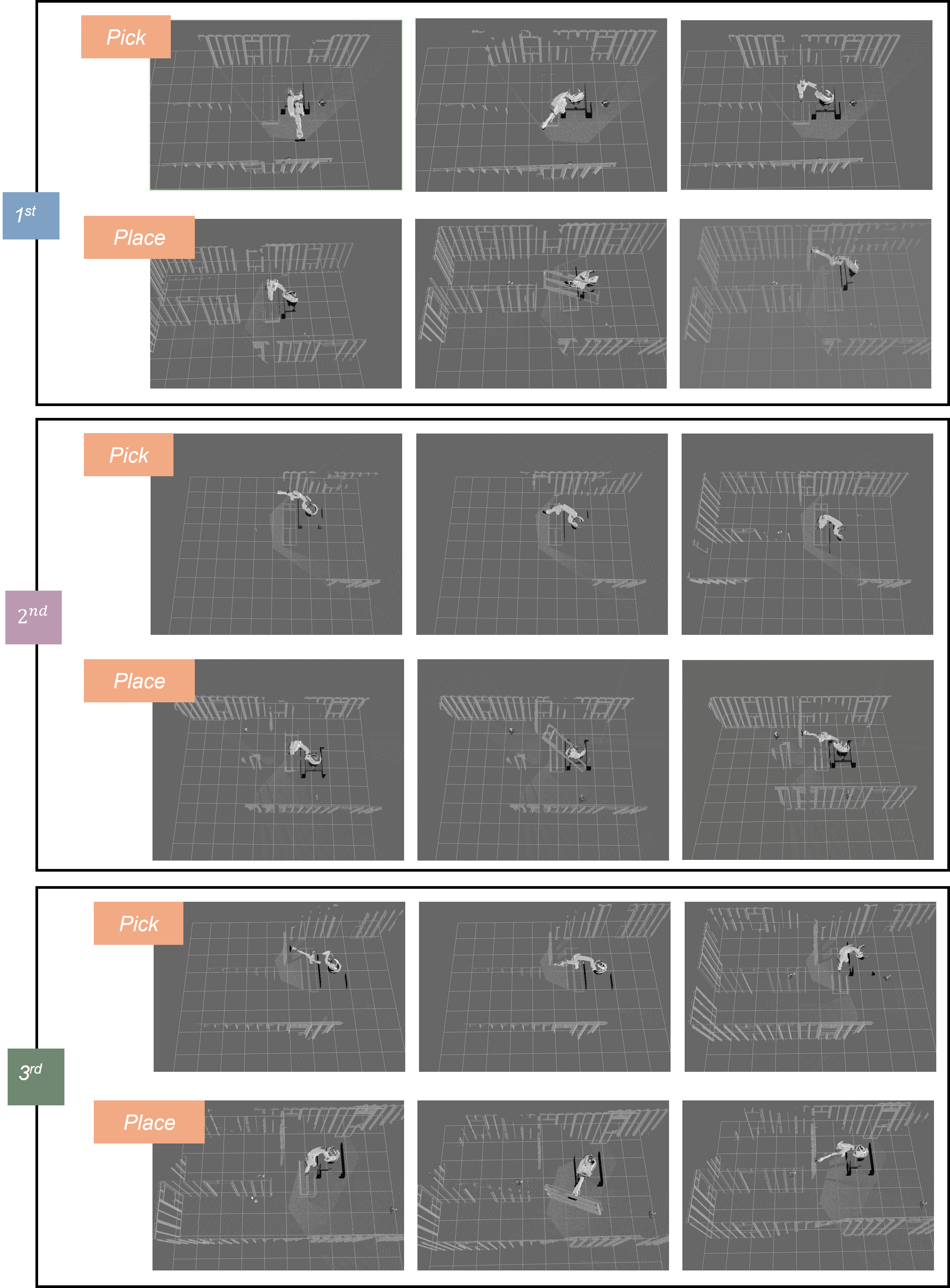}
  \caption{Point cloud generated by depth cameras on supervising robot}
  \label{fig:case-study-pcds}
\end{figure*}

Once the viewpoints are selected, the supervising robots are controlled to position their cameras at the target viewpoints. After the supervising robots are in placing, the construction robot executes its planned trajectory. This process is repeated for each target frame installation: generating a picking trajectory, repositioning the supervising robots, performing the picking operation; then generating a placing trajectory, repositioning, and performing the placement. Figure \ref{fig:case-study-images} illustrates the entire process, displaying real RGB images captured by supervising robots' depth cameras during each trajectory execution, along with the snapshots of the simulated Gazebo environment at the same moment. The depth data from the cameras is used to generate the point cloud, as shown in Figure \ref{fig:case-study-pcds}, which is created by registering the depth data the supervising robots' cameras.

\begin{table*}[h!]
  \centering
  \caption{Summary of distance and coverage/visibility}
    \begin{scriptsize}
    \begin{tabularx}{0.8\textwidth}{c *{3}{X} c *{6}{X}}
      \hline
      \textbf{Target} & \multicolumn{3}{c}{\textbf{Pick}} & & \multicolumn{6}{c}{\textbf{Place}} \\ 
      \cline{2-4} \cline{6-11}
      & $C(V)$ & $d(G(s), v_1)$ & $d(G(s), v_2)$ & & $C(V)$ & $d(G(s), v_1)$ & $d(G(s), v_2)$ & $d(\text{obj}, v_1)$ & $d(\text{obj}, v_2)$ & $AvgVis(V)$ \\ 
      \hline
      1\textsuperscript{st} & 1.00 & 1.74 & 2.32 & & 0.998 & 2.36 & 3.56 & 3.19 & 3.52 & 0.76 \\ 
      2\textsuperscript{nd} & 1.00 & 1.75 & 2.69 & & 0.999 & 2.29 & 3.51 & 3.28 & 3.18 & 0.70 \\ 
      3\textsuperscript{rd} & 1.00 & 1.97 & 2.47 & & 1.00  & 2.89 & 3.29 & 3.09 & 3.39 & 0.80 \\ 
      \hline
    \end{tabularx}
    \end{scriptsize}
    \label{tab:summary}
\end{table*}

Table \ref{tab:summary} details the viewpoint selection results for each trajectory during the frame installation process. For each trajectory, the coverage $C(V)$ is calculated using Eq.~\ref{eq:CVm}, which assesses the spatial coverage of the construction robot's motion envelope along the trajectory. The distances $d\left(G\left(s\right),v_1\right)$ and $d\left(G\left(s\right),v_2\right)$, defined in Eq.~\ref{eq:eq8}, quantify the distances between the robot's motion envelope center and the two selected viewpoints. For the placing trajectories, in addition to coverage $C(V)$ and distances $d\left(G\left(s\right),v_1\right)$ and $d\left(G\left(s\right),v_2\right)$, the visibility of the target object over the trajectory $s$ under the selected viewpoint combination $AvgVis(V)$ is computed using Eq.~\ref{eq:AvgVis}. The average distance from the target object to the viewpoints $d\left(P_{obj},v_1\right)$ and $d\left(P_{obj},v_2\right)$ across the trajectory are calculated using Eq.~\ref{eq:eq9}.

The viewpoint selection method for the supervising robots demonstrated strong performance. For all the picking trajectories, full coverage (1.00) is achieved, with the viewpoints positioned at distances within 2.69 meters from the center of the robot. For all the placing trajectories, the spatial coverage of the robot’s motion envelope exceeded 0.998, ensuring that the robot's movements are mostly within view. Meanwhile, the minimum visibility of the target object is 0.70, allowing the majority of the target object to be observed throughout the process. The distances from the viewpoints to the robot's motion envelope center are below 3.56 meters, and the average distances to the target object are below 3.52 meters.

\section{Experimental Validation of Viewpoint Selection Module}\label{experimental-validation}

To validate the robustness of the Viewpoint Selection Module under various constraints and complexities, a series of simulation experiments were conducted. These experiments include variations in space constraints, material positioning, and environmental complexity, designed to mirror real-world construction challenges and to test the flexibility of the viewpoint selection method under different conditions. Space constraints restrict the operational boundaries of all robots in the team, directly affecting the spatial range of viable viewpoints. Material positioning affects the distance and manipulation trajectory from the material to the installation location. Greater distances result in longer potential end-effector trajectories, possibly leading to a larger robot motion envelope. Environmental complexity pertains to the built environment's complexity, with more partitions and complex layouts posing higher challenges for viewpoint selection due to increased occlusion and higher potential for collisions. The hardware setup, including the construction robot, supervising robots, and depth cameras, is consistent with that of the Case Study. In each experiment, the construction robot is tasked with picking up one prefabricated frame and installing it at the target position. Quantitative metrics for evaluating the objectives (i.e., coverage, distance, and visibility) of the selected viewpoints are provided and discussed.

\subsection{Impact of Space Constraints}

\begin{figure}[h!]
  \centering
  \includegraphics[width=\dimexpr\columnwidth]{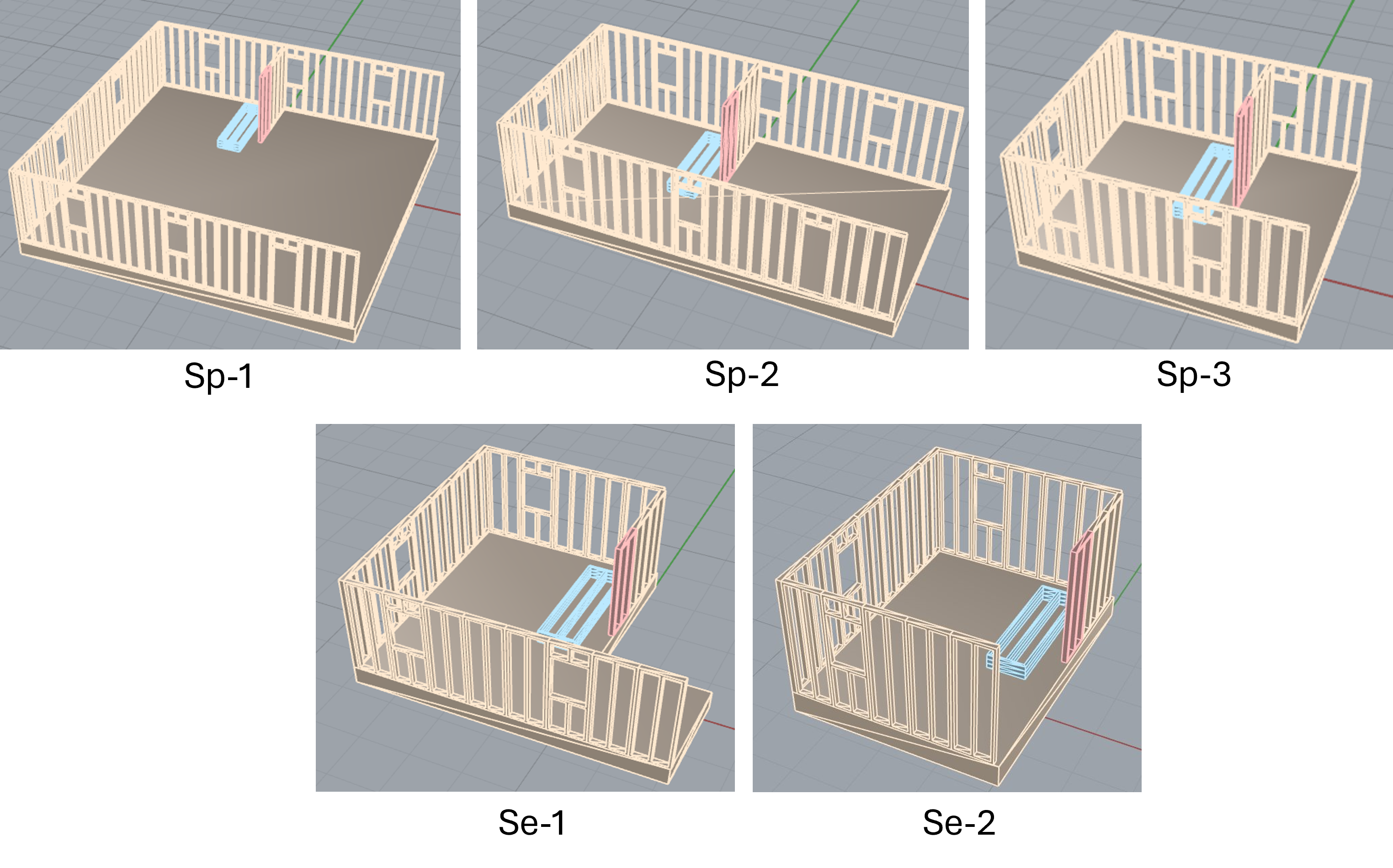}
  \caption{BIMs for different space constraints}
  \label{fig:space_constraints}
\end{figure}

This section explores how variations in operable floor area impact viewpoint selection. Figure \ref{fig:space_constraints} shows the BIMs under five environmental settings: the first three (Sp-1, Sp-2, and Sp-3) focus on partition wall frame installation, while the last two (Se-1 and Se-2) involve exterior wall frame installation. Partition wall frame installations offer greater observation flexibility from both sides of the frame, whereas exterior wall frame installations can only be observed from the indoor side.

\begin{figure*}[h!]
  \centering
  \includegraphics[width=1\textwidth]{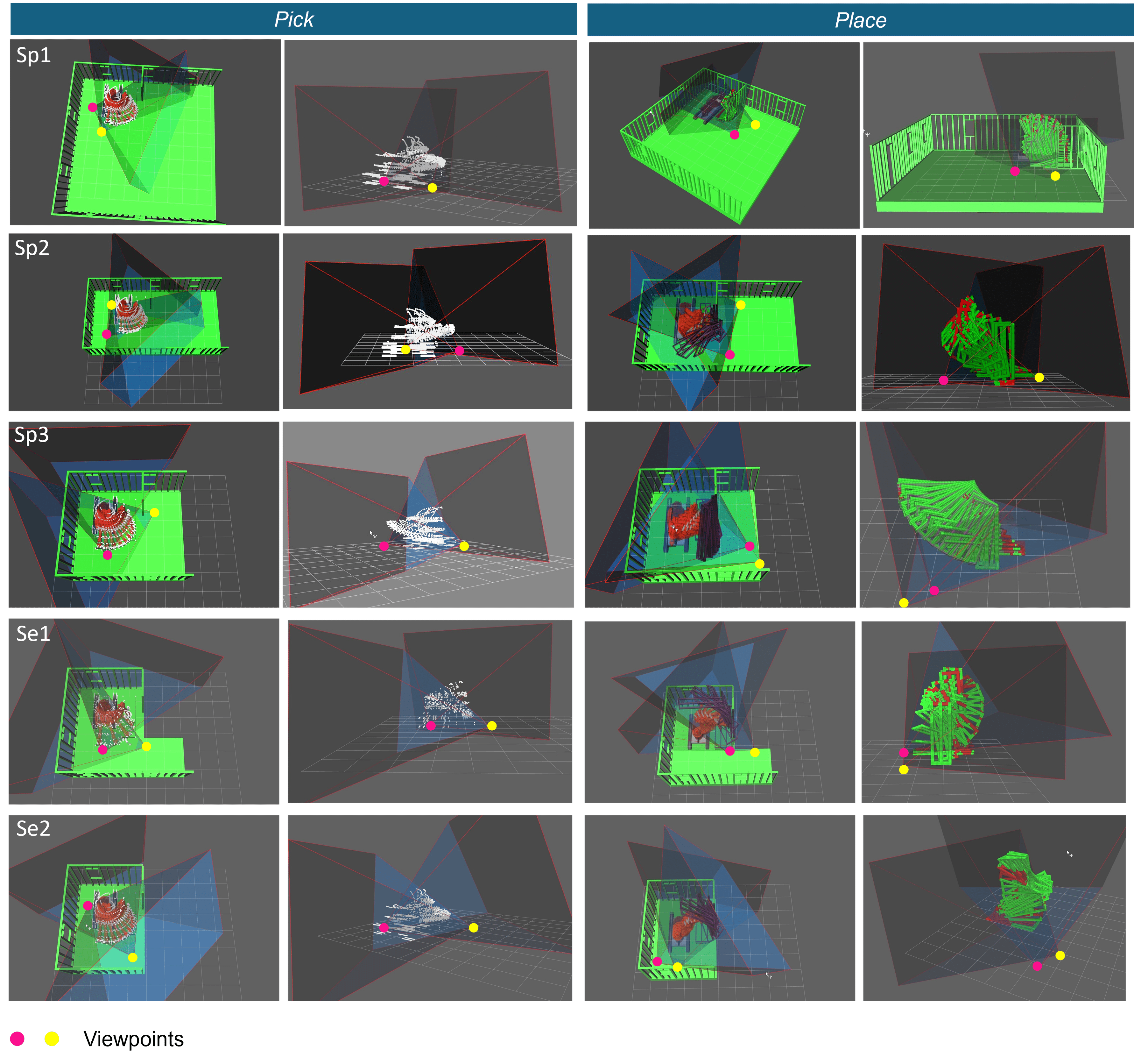}
  \caption{Viewpoint selection results for different space constraints}
  \label{fig:viewpoint-space-constraints}
\end{figure*}

Figure~\ref{fig:viewpoint-space-constraints} presents the viewpoint selection results for each scenario, detailing the camera positions, coverage of robot's motion envelope, and visibility of the target object. The point cloud representation of the picking motion envelope is displayed as white points, with the view cones represented by dark blue transparent cones. For the placing trajectories, the last row presents the overall visibility result for points on the target object along the entire trajectory. Visible points from the selected viewpoint combination are shown in green, while invisible points are marked in red. Quantitative metrics for distance, coverage, and visibility are provided in Table ~\ref{tab:space-constraints}. Despite differences in space constraints, the selected viewpoints consistently achieved full coverage for all picking trajectories. For placing trajectories, the coverage remains complete for the two relatively larger spaces, Sp-1 and Sp-2. For exterior wall frame installation cases (Se-1 and Se-2), coverage drops further to 0.995 and 0.986, respectively, due to restricted observation access. In terms of distance, it is found that the average distance between cameras and the target object is the highest in Se-2, the most constrained space. Visibility of the target object ranges between 0.73 and 0.91 across the five scenarios, with the lowest value of 0.73 observed in Sp-2, in which the robot trajectory involves detours, leading to a more dispersed distribution of the target object along the trajectory.

 \begin{table*}[h!]
  \centering
  \caption{Space constraints experiment results}
  \begin{scriptsize}
  \begin{tabularx}{\textwidth}{c *{3}{X} c *{6}{X}}
    \hline
    \textbf{Scenario} & \multicolumn{3}{c}{\textbf{Pick}} & & \multicolumn{6}{c}{\textbf{Place}} \\ 
    \cline{2-4} \cline{6-11}
    & $C(V)$ & $d(G(s), v_1)$ & $d(G(s), v_2)$ & & $C(V)$ & $d(G(s), v_1)$ & $d(G(s), v_2)$ & $d(\text{obj}, v_1)$ & $d(\text{obj}, v_2)$ & $AvgVis(V)$\\ 
    \hline
    Sp-1 & 1.00 & 1.99 & 2.37 & & 1.00 & 3.34 & 3.61 & 2.68 & 2.72 & 0.86\\
    Sp-2 & 1.00 & 1.69 & 2.45 & & 1.00  & 3.20 & 3.53 & 2.75 & 3.26 & 0.73 \\
    Sp-3 & 1.00 & 1.92 & 2.46 & & 0.998 & 3.29 & 4.12 & 2.66 & 3.47 & 0.91 \\
    Se-1 & 1.00 & 2.07 & 2.51 & & 0.986 & 2.23 & 3.28 & 2.46 & 2.99 & 0.80 \\
    Se-2 & 1.00 & 1.75 & 2.57 & & 0.995 & 2.82 & 3.11 & 3.73 & 4.23 & 0.76\\
    \hline
  \end{tabularx}
  \end{scriptsize}
  \label{tab:space-constraints}
\end{table*}

\subsection{Impact of Material Positioning}

\begin{figure}[h!]
  \centering
  \includegraphics[width=\dimexpr\columnwidth]{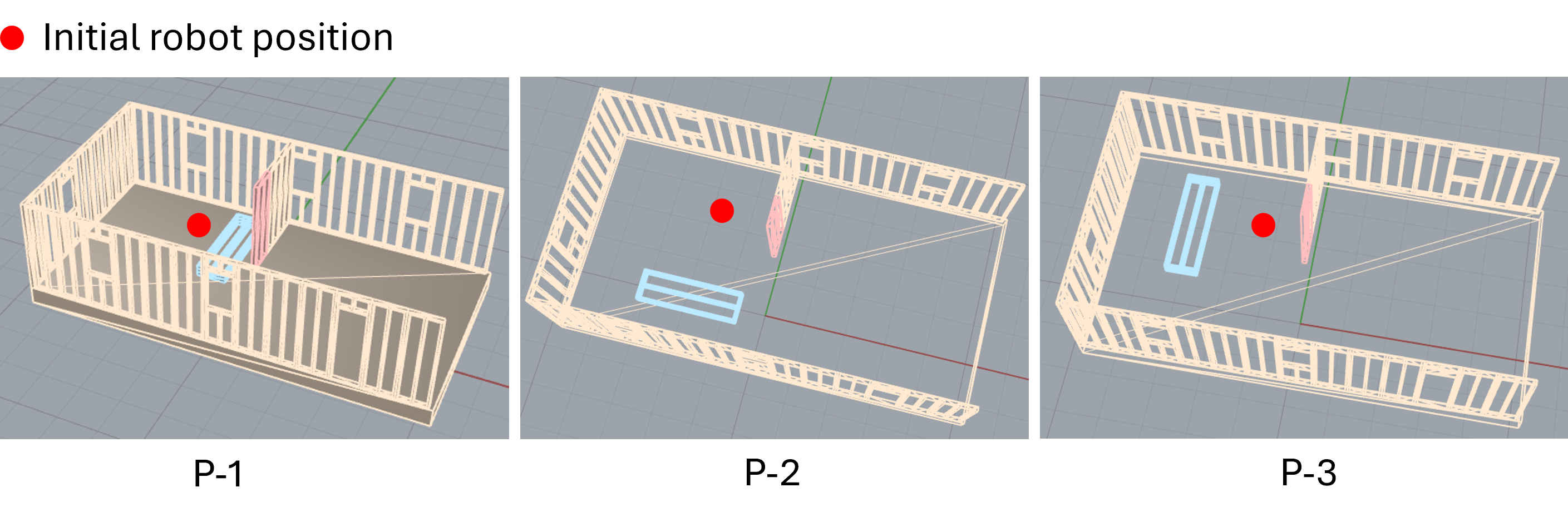}
  \caption{BIMs for material positioning experiments} 
  \label{fig:bim-material-positioning}
\end{figure}

\begin{figure*}[h!]
  \centering
  \includegraphics[width=1.0\textwidth]{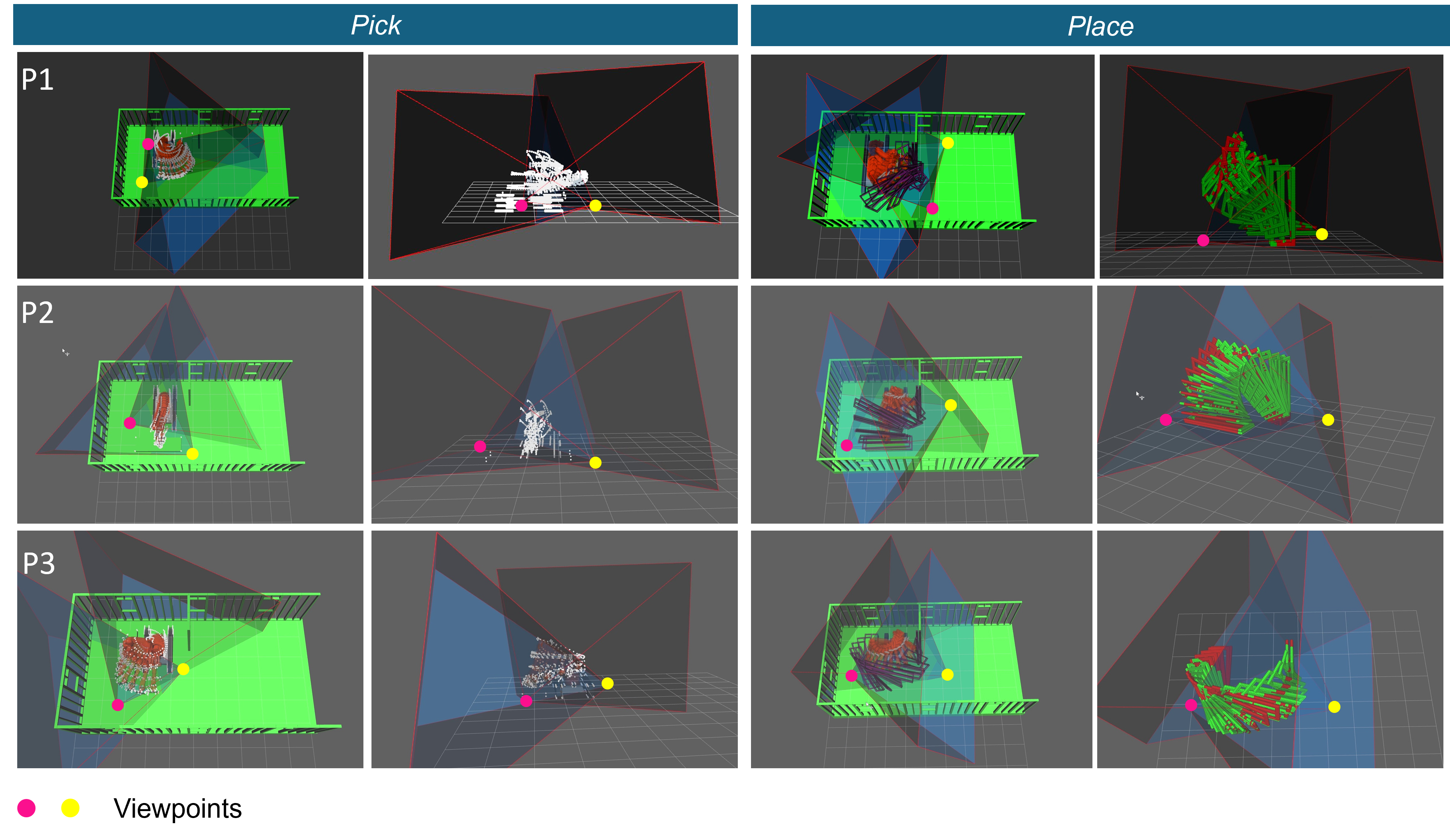}
  \caption{Viewpoint selection results of different material positions}
  \label{fig:viewpoints-material-positioning}
\end{figure*}

Varying the placement of materials affects both the length of the generated trajectories and the spatial footprint of the robot to execute the trajectory. This section evaluates the robustness of the proposed viewpoint selection method under varying material positioning. Figure \ref{fig:bim-material-positioning} illustrates three material positioning settings (P-1, P-2, and P-3), all with the same robot initial configuration and environment setup. The material is placed progressively farther from the target location: P-1 is closest to the target\, P-2 is further away, and P-3 is on the other side of the robot. The viewpoint selection results for these scenarios are illustrated in Figure \ref{fig:viewpoints-material-positioning}, with the corresponding metrics summarized in Table ~\ref{tab:material-pos}. For the picking trajectories, despite variations in material positioning, the coverage consistently achieves 1.00, with the farthest average distance between the camera and the robot motion envelope center recorded  as 2.45 meters. For the placing trajectories, the coverage still achieves 1.00, with the visibility metrics for P-1, P-2, and P-3 being 0.73, 0.72, and 0.74, respectively.

\begin{table*}[h!]
  \centering
  \caption{Material positioning experiment results}
  \begin{scriptsize}
  \begin{tabularx}{\textwidth}{c *{3}{X} c *{6}{X}}
    \hline
    \textbf{Target} & \multicolumn{3}{c}{\textbf{Pick}} & & \multicolumn{6}{c}{\textbf{Place}} \\ 
    \cline{2-4} \cline{6-11}
    & $C(V)$ & $d(G(s), v_1)$ & $d(G(s), v_2)$ & & $C(V)$ & $d(G(s), v_1)$ & $d(G(s), v_2)$ & $d(\text{obj}, v_1)$ & $d(\text{obj}, v_2)$ & $AvgVis(V)$ \\ 
    \hline
    P-1 & 1.00 & 1.69 & 2.45 & & 1.00 & 3.20 & 3.53 & 2.75 & 3.26 & 0.73 \\
    P-2 & 1.00 & 2.02 & 2.51 & & 1.00 & 3.15 & 3.35 & 3.39 & 3.38 & 0.72 \\
    P-3 & 1.00 & 1.79 & 2.41 & & 1.00 & 2.59 & 3.32 & 2.59 & 3.46 & 0.74 \\
    \hline
  \end{tabularx}
  \end{scriptsize}
  \label{tab:material-pos}

\end{table*}

\subsection{Impact of Environmental Complexity}\label{env-complexity}

\begin{figure}[h!]
  \centering
  \includegraphics[width=\dimexpr\columnwidth]{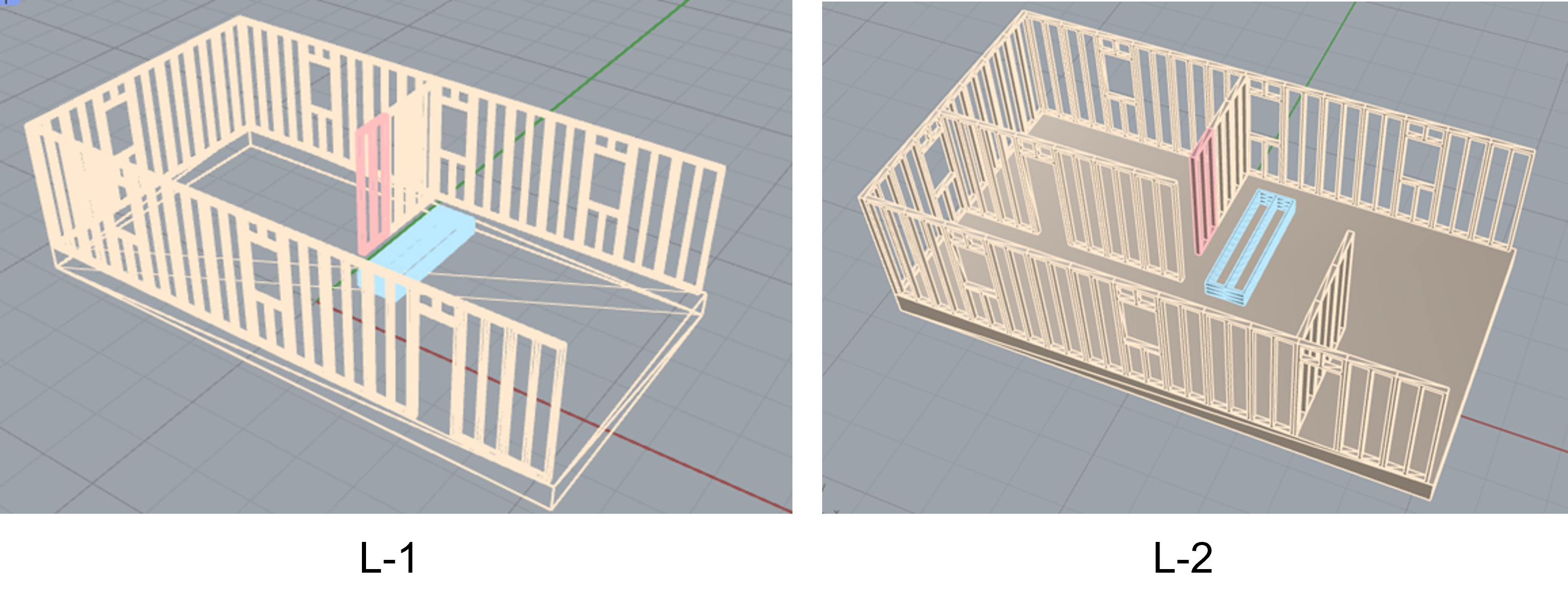}
  \caption{BIMs for environmental complexity experiments}
  \label{fig:BIM-env}
\end{figure}

\begin{figure*}[h!]
  \centering
  \includegraphics[width=1.0\textwidth]{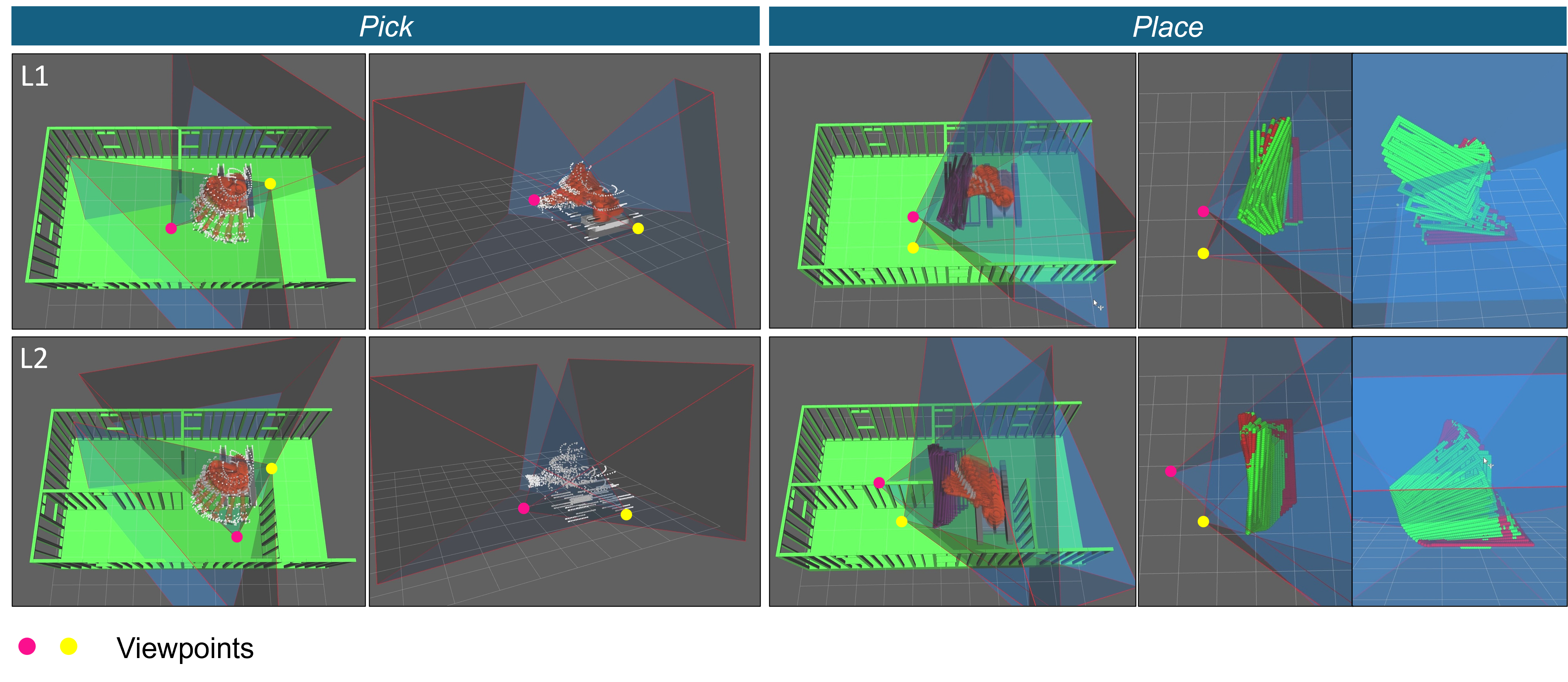}
  \caption{Viewpoint selection results of different layout complexity}
  \label{fig:viewpoints-layout}
\end{figure*}

Two scenarios with varying environmental complexity are designed (Figure \ref{fig:BIM-env}). Compared with L-1, L-2 introduces additional partition frames, increasing the potential for occlusions. The viewpoint selection results for L-1 and L-2 are shown in Figure~\ref{fig:viewpoints-layout}, with the associated metrics summarized in Table ~\ref{tab:env-complexity}. For the picking trajectories,  both scenarios achieve full coverage with the maximum average distance of 2.28 meters. For the placing trajectories, the coverage achieved in L-2 drops slightly to 0.996, compared to 1.00 in L-1. Additionally, visibility also declines in  L-2, dropping to 0.81 from 0.86 in L-1. These reductions in coverage and visibility are likely due to the additional partition frames in L-2, which introduce occlusions that limit the supervising robot’s ability to maintain a clear line of sight to the target object during placement. 

\begin{table*}[h!]
  \centering
  \caption{Environmental complexity experiment results}
  \begin{scriptsize}
  \begin{tabularx}{\textwidth}{c *{3}{X} c *{6}{X}}
    \hline
    \textbf{Target} & \multicolumn{3}{c}{\textbf{Pick}} & & \multicolumn{6}{c}{\textbf{Place}} \\ 
    \cline{2-4} \cline{6-11}
    & $C(V)$ & $d(G(s), v_1)$ & $d(G(s), v_2)$ & & $C(V)$ & $d(G(s), v_1)$ & $d(G(s), v_2)$ & $d(\text{obj}, v_1)$ & $d(\text{obj}, v_2)$ & $AvgVis(V)$\\ 
    \hline
    L-1 & 1.00 & 2.28 & 2.09 & & 1.00 & 2.79 & 3.32 & 2.11 & 2.70 & 0.86 \\
    L-2 & 1.00 & 2.08 & 2.17 & & 0.996 & 2.78 & 3.57 & 2.31 & 2.72 & 0.81\\
    \hline
  \end{tabularx}
  \end{scriptsize}
  \label{tab:env-complexity}

\end{table*}

\section{Discussion}\label{discussion}

As a proof-of-concept implementation, a case study involving the installation of three prefabricated wall frames was conducted. The case study demonstrates the feasibility of the proposed framework, where supervising robots collaborate with the construction robot by positioning their cameras at strategically selected viewpoints to monitor each picking or placing operation during the assembly process. The number of supervising robots is capped at two, allowing the viewpoint selection module to choose either a single viewpoint using one supervising robot or a combination of two viewpoints that requires two robots. Notably, the two-viewpoint combination is always selected over the single-viewpoint solution. This is likely because the two-viewpoint combination provides better coverage and distance balance.  Additionally, the redundancy introduced by the two-viewpoint combinations enhance the system's robustness against occlusions. 


In addition, for some picking processes, although full coverage is achieved, tracks of the construction robot mobile base sometimes occupy a significant portion of the visual capture (Figure ~\ref{fig:case-study-images}). This issue arises because the viewpoint determination for the picking process only considers coverage and proximity but does not account for visibility, as no target object is involved. This finding highlights the importance of considering target object visibility in viewpoint determination.

A series of experiments were conducted to further evaluate the performance viewpoint selection method. Consistent with the observation in the case study, all experiments chose a combination of two cameras over a single camera. For picking trajectories, the spatial coverage of the selected viewpoints consistently reached 1.00 (fully covered), with a maximum distance of 2.51 meters, which is less than the 10-meter maximum depth range of the depth camera, ensuring that the captured data maintains an acceptable quality. For placing trajectories, the viewpoint selection method also demonstrated robust performance, with the coverage range from 0.986 to 1 and the visibility of the target object ranging from 0.72 to 0.91. 

The experiments reveal several important insights about the factors that influence viewpoint selection performance. First, the results show that the planned trajectory significantly affects visibility. When the planned trajectory is extended, the target object becomes more spatially dispersed as it moves through an extended path (e.g., P-1 (Sp-2), P-2, P-3). This reduces visibility to below 0.80, highlighting challenges in capturing dispersed targets. Additionally, visibility is similar across different material positioning settings despite the differences in the distance between the material and the target installation location, further indicating that target visibility is heavily influenced by the planned trajectory. Second, space constraints also have a notable impact on visibility. The largest space setting, Sp-1, achieves higher visibility due to the availability of more favorable viewpoints. In constrained spaces such as Se-2, visibility drops to 0.76 as the supervising robots struggle to access optimal viewpoints. Additionally, the complexity of the environment further influences the selection results. In complex layouts, such as L-2 (with partitions), visibility decreases from 0.86 (L-1) to 0.81 since the partition obstructed the line of sight to the target object.  

\begin{figure}[h!]
  \centering
  \includegraphics[width=\dimexpr\columnwidth]{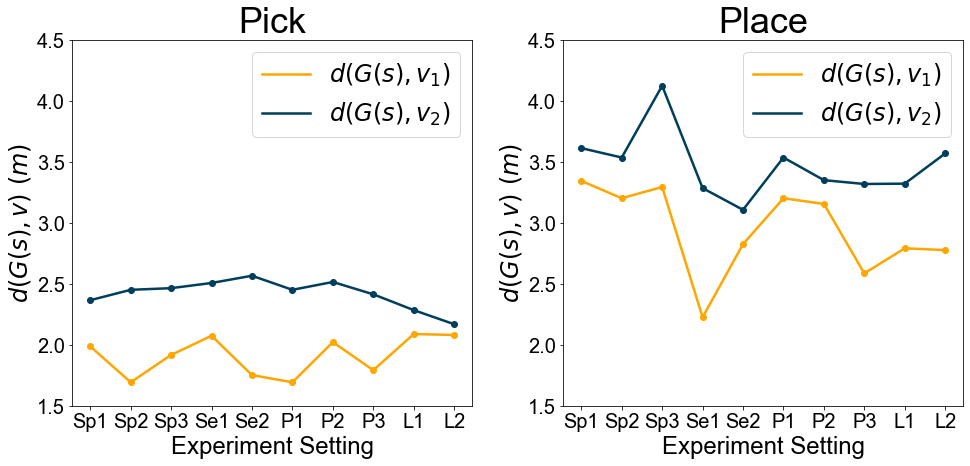}
  \caption{Distances to robot motion envelope center under different experiment settings}
  \label{fig:exp_distance}
\end{figure}

\begin{figure}[h!]
  \centering
  \includegraphics[width=\dimexpr\columnwidth]{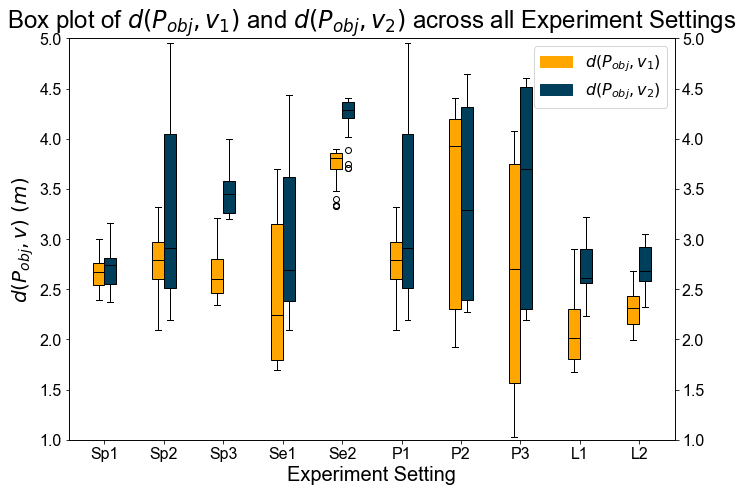}
  \caption{Box plot of distances to target object during each operation}
  \label{fig:exp_boxplot}
\end{figure}

The distance results further demonstrate the impact of trajectory planning. Compared to the distances to the robot's motion envelope center for picking operations, those for placing operations are generally larger (Figure~\ref{fig:exp_distance}). This difference is attributed to the larger spatial area occupied by the construction robot handling large frames compared to operating without payload. As a result, to capture the entire motion envelope, including the robot and the frame, the supervising robots must position themselves further away. The Figure~\ref{fig:exp_boxplot} shows the distribution of the distance between the camera and the target object center at different joint states along the placing trajectory. It can be observed that the distance to the target object exhibits significant variability in most experiment settings. However, in shorter trajectories, such as those from Sp-1, Se-2, and L-2, the distances show lower variance across different joint states because the target object's movement is more confined, resulting in closer target centers. This observation highlights two key insights: (1) the distance variance results heavily depend on the generated trajectory, emphasizing the critical role of optimized trajectory planning; and (2) selecting viewpoints that minimize not only the distance average but also variance is critical to maintaining consistent capture performance throughout the operation. In the future, trajectory planning will be optimized to minimize detours, create more direct paths. Additionally, the viewpoint selection method will be further refined to consider the distance variance as an additional objective in the optimization process.

Beyond the specific observations, the proposed framework itself demonstrates scalability, adaptability, and generalizability, making it a versatile solution for various robotic applications. The proposed framework is scalable to a higher number of supervising robots by extending the chromosome of the individual in NSGA-II, as detailed in Section \ref{combination-of-camera-poses}, thereby providing more comprehensive visual information. It is also adaptable to different tasks by adjusting the target object and specific coverage area. For example, in the case of robots welding steel structures, the end-effector could be designated as the target object, with the coverage area minimized to the end-effector’s motion envelope and the welding space. Furthermore, the framework is generalizable to different types of robots, such as combinations of mobile manipulators and drones, as the viewpoints are sampled in the configuration space of each robot type. The proposed multi-robot system enables supervising robots to provide rich 2D RGB visual data (Figure~\ref{fig:case-study-images}) and 3D spatial information of the construction process (Figure~\ref{fig:case-study-pcds}). This information can be further processed to enhance the construction robots’ understanding of their surroundings.

However, there are still some limitations in this paper. First, the proposed approach relies on as-built BIMs that accurately reflect actual site conditions, requiring techniques such as Scan-to-BIM \cite{chen2022rapid, wang2024omni} and change detection\cite{chuang2023, meyer2022} as a prior step. Second, this paper does not account for the dynamic obstacles (e.g., human workers) that intrude into the workspace during the robot execution. Future work will integrate visual data captured by the supervising robots with the construction robot decision-making process for the robot to dynamically adapt to site changes, and adaptive motion planning and supervising robot control techniques will be explored. Additionally, while multiple supervising robots could provide a better trade-off of coverage, proximity, and visibility, each view only contains partial information. This results in fragmented visual data, particularly for RGB images. Future work could address this limitation by using computer vision techniques, such as image stitching, or deploying an additional supervising robot dedicated to providing a comprehensive view. Lastly, optimization could be enhanced by incorporating more objectives, such as supervising robot travel costs and target object distance variance.

\section{Conclusion}\label{conclusion}

Visual perception data is critical for adaptive control and responsive behavior in robotic operations, particularly in unstructured construction sites. However, the provision of task-relevant visual information during construction robot operation has often been overlooked. To address this gap, this study proposes a multi-robot coordination framework to enhance the visual perception of the robotic construction process. The framework integrates agile supervising robots, such as mobile bases with lightweight manipulators, to coordinate with the primary construction robot by collecting visual information tailored to the upcoming construction robot operation. A viewpoint selection method is proposed to determine the positioning of supervising robots, maximizing coverage and proximity while incorporating visibility considerations. This method represents the boundary of construction robot's motion envelope as a point cloud and each joint state representation along the trajectory as a 3D OGM, enabling accurate and efficient viewpoint assessment. By sampling camera poses within the supervising robot's configuration space, the framework eliminates the need for computationally expensive inverse kinematics. The viewpoint is determined by optimizing coverage and proximity, while considering visibility. First, NSGA-II-based optimization balances proximity and coverage, forming a set of candidate results. Subsequently, the visibility of these candidates is evaluated, and the one with the highest visibility is selected. This approach significantly reduces the computational burden of visibility evaluation, thereby enhancing overall efficiency. The proposed framework is validated through a case study and a series of simulation experiments, demonstrating robust performance under various constraints and complexities. The task-specific visual data provided by the proposed framework can be further used for computer vision analyses, enabling robots to interpret and adapt to dynamic environments and facilitating human-robot interaction.

\bibliographystyle{elsarticle-num} 
\bibliography{journal1.bib}

\end{document}